\documentclass[a4paper,conference]{IEEEtran}
\usepackage{cite}
\usepackage{amsmath}

\usepackage[utf8]{inputenc}

\usepackage[inline]{enumitem}
\usepackage{array}

\usepackage{tikz}
\usepackage{pgfplots}
\input{tikz-pgfplots.settings}

\def\thicknesslineplot{ultra thick}
\def\thicknesshack{0.085cm}
\def\mythickness{ultra thick}
\def\thicknesslineplotTwo{thick}
\def\mythicknesssmall{thick}
\def\thicknesslineplotThree{very thick}

\definecolor{gtcolor}{rgb}{0,0,0}
\definecolor{ourcolor}{rgb}{0,0.4,0.6}
\definecolor{dingcolor}{rgb}{0.8,0.1,0.3}
\definecolor{kukelovacolor}{rgb}{0,0.6,0}
\definecolor{bougnouxcolor}{rgb}{0,0.7,0.7}
\definecolor{inliercolor}{rgb}{0.2,1,0.1}
\definecolor{outliercolor}{rgb}{1,0,0}
\definecolor{smallcolor1}{rgb}{0.,0,0}
\definecolor{smallcolor2}{rgb}{0,0,0}

\usepackage{standalone}

\usepackage{bm}
\renewcommand{\vec}[1]{\bm{#1}}
\providecommand{\mat}[1]{\bm{#1}}

\usepackage{mathtools}

\newcommand{\te}[1]{\text{#1}}
\DeclarePairedDelimiterX{\norm}[1]{\lVert}{\rVert}{#1}
\DeclarePairedDelimiterX{\abs}[1]{\lvert}{\rvert}{#1}
\newcommand{\fr}[2]{\frac{#1}{#2}}

\newcommand{\T}{T}
\DeclareMathOperator{\tr}{tr}

\DeclareMathOperator{\diag}{diag}
\makeatletter
\newcommand{\vast}{\bBigg@{4}}
\newcommand{\Vast}{\bBigg@{5}}
\makeatother

\newcommand{\etal}{et~al.}
\newcommand{\ie}{i.e.}
\newcommand{\eg}{e.g.}

\newcommand{\Hc}{\ensuremath{\mat{H}_{\mathrm{euc}}}}
\newcommand{\Hy}{\ensuremath{\mat{H}_{y}}}
\graphicspath{{images/}}

\usepackage{hyperref}
\usepackage[capitalize]{cleveref}
\begin{document}
%
\title{Minimal Solvers for Indoor UAV Positioning}



%
\author{\IEEEauthorblockN{
Marcus Valtonen \"{O}rnhag\IEEEauthorrefmark{1},
Patrik Persson\IEEEauthorrefmark{1},
M{\aa}rten Wadenb\"{a}ck\IEEEauthorrefmark{2},
Kalle {\AA}str{\"o}m\IEEEauthorrefmark{1},
Anders Heyden\IEEEauthorrefmark{1}
}

\IEEEauthorblockA{\IEEEauthorrefmark{1}Centre for Mathematical Sciences\\
Lund University\\
Lund, Sweden
}
\IEEEauthorblockA{\IEEEauthorrefmark{2}Department of Electrical Engineering\\
Link\"{o}ping University\\
Link\"{o}ping, Sweden
}
Email: marcus.valtonen\_ornhag@math.lth.se
}


\maketitle

\begin{abstract}
In this paper we consider a collection of relative pose problems which arise naturally in applications for visual indoor UAV navigation.
We focus on cases where additional information from an onboard IMU is available and thus provides a partial extrinsic calibration through the gravitational vector.
The solvers are designed for a partially calibrated camera, for a variety of
realistic indoor scenarios, which makes it possible to navigate using images of the ground floor.
Current state-of-the-art solvers use more general assumptions, such as using arbitrary
planar structures; however, these solvers do not yield adequate reconstructions for real
scenes, nor do they perform fast enough to be incorporated in real-time systems.

We show that the proposed solvers enjoy better numerical stability, are faster, and require
fewer point correspondences, compared to state-of-the-art solvers. These properties are vital components for
robust navigation in real-time systems, and we demonstrate on both synthetic and real
data that our method outperforms other methods, and yields superior motion
estimation\footnote{Code available
at:~\href{https://github.com/marcusvaltonen/minimal_indoor_uav}{https://github.com/marcusvaltonen/minimal\_indoor\_uav}.}.
\end{abstract}


%
\IEEEpeerreviewmaketitle

\section{Introduction}
One of the lessons which have been learned in computer vision is the importance of leveraging prior knowledge pertaining to the specific vision task at hand.
For geometrical computer vision problems, this often means introducing constraints encoding the prior knowledge already at the modeling stage.
Successfully making use of such prior information in a vision system can have numerous direct benefits, such as improved robustness, better accuracy, and faster performance.
However, constraining the solution space will in many cases come at the cost of requiring significantly more complex algorithms.
A familiar illustration of this is the computation of fundamental or essential matrices in epipolar geometry; the simplicity of the seven-point~\cite{hartley-pami-1994} or eight-point~\cite{longuet-higgins-nature-1981} algorithms stands in stark contrast to the complexity of the five-point algorithm~\cite{nister-pami-2004}.
\begin{figure}[t!]
\centering
\includegraphics[width=\linewidth]{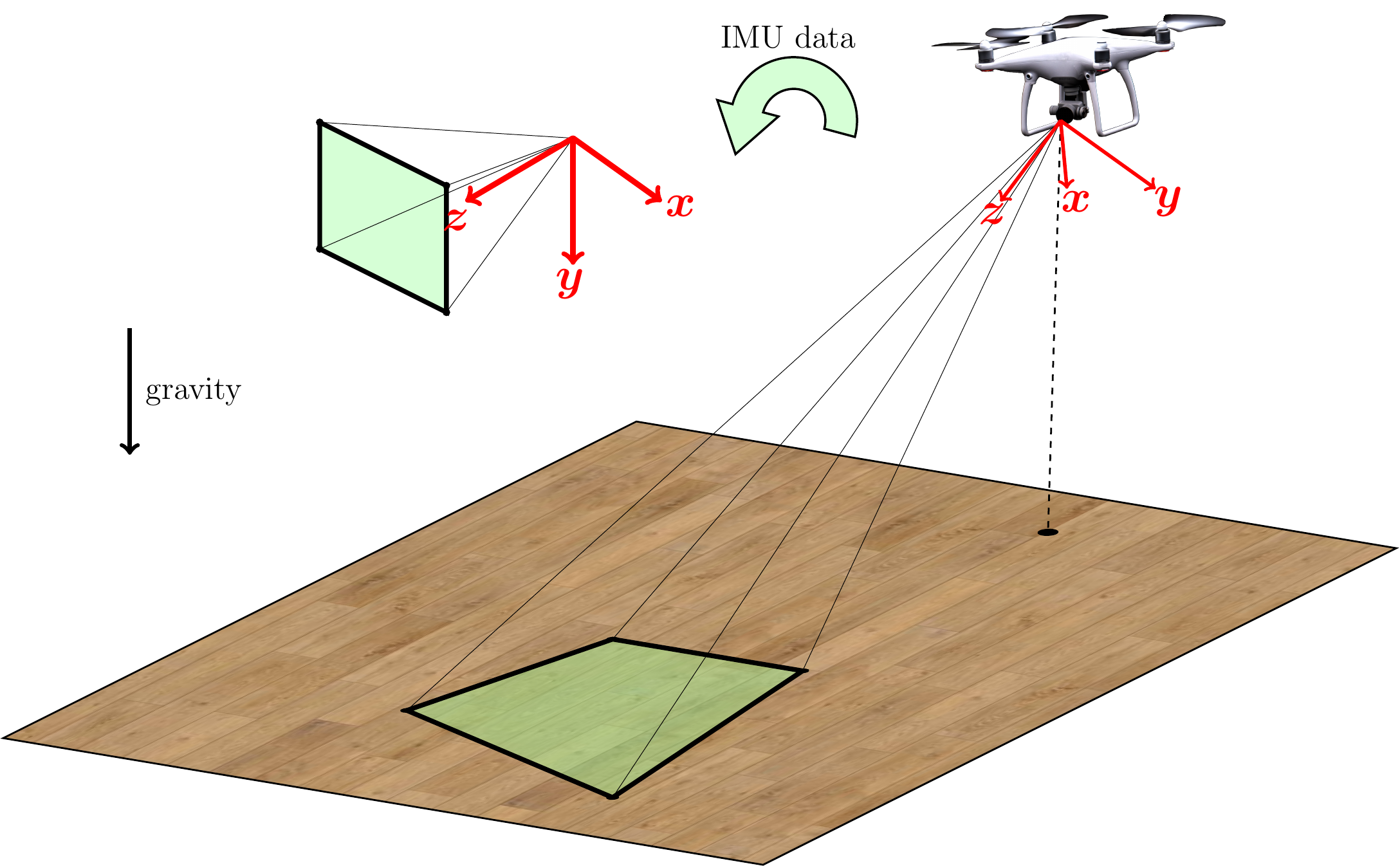}
\caption{The pitch and roll angles can be estimated from the IMU data, leaving the unknown
yaw angle to be determined.}
\label{fig:1}
\end{figure}

In particular over the past two decades, the embracement of approaches based on algebraic geometry has allowed many constrained geometrical problems to be solved with a minimal (or near-minimal) amount of data.
This has notably been the case for notorious problems such as five-point essential matrix estimation~\cite{nister-pami-2004,li-hartley-icpr-2006,stewenius-etal-ijprs-2006} or optimal three-view triangulation~\cite{stewenius-iccv-2005,byrod-etal-accv-2007,kukelova-etal-2013}, but also for pose estimation problems involving some kind of geometric distortion such as radial distortion~\cite{josephson-byrod-cvpr-2009,kukelova-etal-cviu-2010,kuang-etal-cvpr-2014,kukelova2015,pritts2017,wu-cvpr-2015,larsson2017ICCVb,larsson2018cvprb} or refractive distortion in underwater problems~\cite{haner-cvpr-2015,palmer-cvpr-2017}.
These successes have been enabled chiefly thanks to improvements in the computational algorithms themselves, but to a lesser extent also by improvements in hardware which have made it feasible to solve increasingly large systems.

Inspired primarily by recent papers by Saurer \etal~\cite{saurer-etal-2017} and by Ding \etal~\cite{ding-etal-2019-iccv}, we would like to create a homography based indoor positioning system for UAVs, that utilizes data from an onboard IMU.
One benefit of using the IMU data is that two degrees of freedom, the roll and the pitch, are removed from the positioning problem, only leaving a single rotational parameter---the yaw angle---to determine.
An illustration of the envisaged situation is provided in \cref{fig:1}.

Our main contributions are threefold:
\begin{enumerate}
    \item We incorporate the known IMU data and a partially calibrated camera with unknown focal length, resulting in new solvers which are orders of magnitude faster than the current state-of-the-art,
    \item We demonstrate through numerical experiments that our solvers are better than or on par with existing solvers with respect to accuracy and runtime, and
    \item We run the solvers in real-time on a UAV system, demonstrating that the derived solvers are feasible for practical real-world situations.
\end{enumerate}

\section{Related Work}
\paragraph{Pose estimation with a known direction}
For many practical pose estimation problems, there are simple ways which allow the extraction of one particular direction, thus giving some of the pose parameters for free.
As shown by Kalantari \etal~\cite{kalantari-etal-jmiv-2011}, this could happen \eg~by detection in the image of a horizon line or vanishing points, or through external sources such as the gravitational vector from an IMU.
Provided a known direction and 2D--3D correspondences, Kukelova \etal~gave a closed-form solution to the absolute pose problem for a calibrated or partially calibrated (unknown focal length) camera~\cite{kukelova-etal-accv-2010}.

Homography-based relative pose, where the plane normal is taken as the known direction, has also been considered in the literature.
Minimal solvers for the calibrated case, including
using the ground floor (2pt) or unknown vertical
plane (2.5pt) and an arbitrary plane (3pt), were treated in~\cite{saurer-iros-2012} and~\cite{saurer-etal-2017}.
In Ding \etal{}~the latter case was considered, and also included
some partially calibrated cases~\cite{ding-etal-2019-iccv}, \eg~with either one or two unknown focal lengths.

\paragraph{Estimation of focal lengths}
Hartley considered the problem of estimating the essential matrix and focal lengths from point correspondences, under the assumption that all other camera parameters were known \cite{hartley1992}.
The method is based on the eight-point algorithm~\cite{longuet-higgins-nature-1981} followed by a series of algebraic manipulations of the fundamental matrix to extract the focal lengths.
Under the same assumptions, Bougnoux derived a simple formula for computing the focal lengths~\cite{bougnoux-iccv-1998}.
With the introduction of polynomial solvers for various problems in computer vision, people have proposed solvers which give directly the essential matrix and the focal lengths, without the necessity to go via the fundamental matrix.
One of the early such algorithms was presented by Li~\cite{li-eccv-2006}, who solved the problem with six point correspondences (the minimal case).
A recent paper by Kukelova \etal~\cite{kukelova-etal-2017-cvpr} contains, as an application, a state-of-the-art solver to the six-point essential matrix and focal length estimation problem.
In the experiments, we will compare our proposed solvers against this last solver by Kukelova~\etal

\paragraph{Solving multivariate polynomial equations}
Solving systems of multivariate polynomial equations numerically is usually done through the \emph{action matrix method}~\cite{moller,kukelova2008,byrod-etal-ijcv-2009}.
This works by first expanding the equations into an \emph{elimination template}, which is then reduced to an eigenvalue problem, where the matrix in question is known as the \emph{action matrix}.
This problem can then be solved efficiently using standard numerical methods.
Depending on how these steps are done, the procedure may or may not introduce spurious solutions which have to be discarded by trial in the original equations.
The sizes of the resulting elimination template and eigenvalue problem will determine how fast the solutions can be computed.

Finding a reasonably sized and numerically stable elimination template for a given problem is a highly non-trivial task, although it has been greatly simplified through the introduction of automatic generators.
One early such automatic generator by Kukelova \etal~\cite{kukelova2008} used a heuristic approach to expand the system of equations until a valid elimination template was obtained, and then proceeded with successively removing one row at a time from the elimination template to prune redundancies.
Larsson \etal~\cite{larsson2017cvpr} exploited the inherent relations between the equations to directly compute a set of monomials with which to expand the equations in order to yield a valid elimination template.

In addition to elimination template generation, there have been several improvements to the numerical accuracy, numerical stability, and speed of polynomial solvers.
In~\cite{eccv/2008/byrod_etal} and \cite{byrod-etal-ijcv-2009}, Byröd \etal\ proposed performing the reduction to the action matrix by means of either a QR decomposition or a singular value decomposition, together with an adaptive scheme for pruning columns which are deemed unnecessary.
Naroditsky and Daniilidis suggested elimination template trimming based on certain algebraic conditions~\cite{naroditsky-iccv-2011}.
However, as it turns out, there are situations where excessive reduction in template size comes at the cost of inferior numerical accuracy~\cite{byrod-thesis}.
Kuang and Åström suggested evaluating many reduction schemes for a large number of random problem instances in order to ensure good numerical properties of the resulting template~\cite{kuang-eccv-2012}.

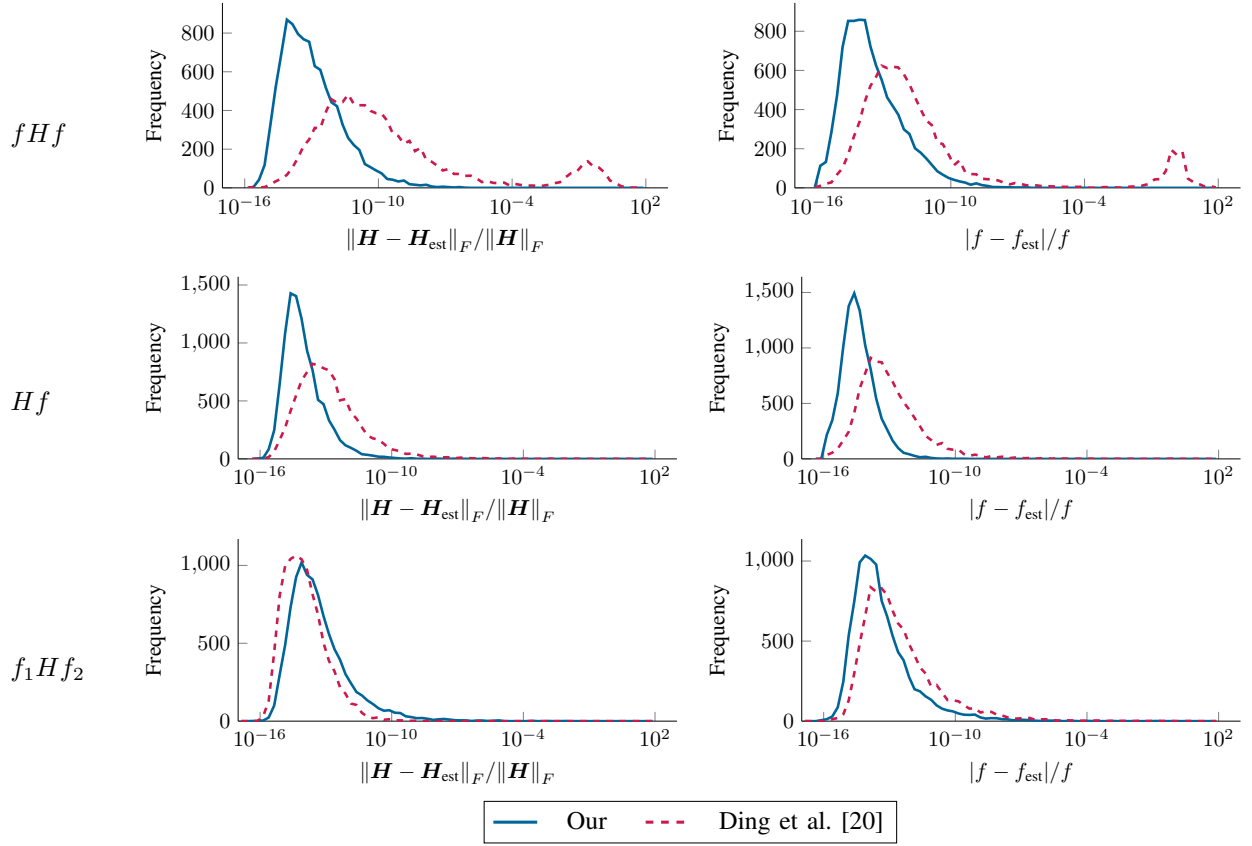
\begin{figure*}[t]
\centering
\begin{tabular}{m{0.06\linewidth} >{\centering\arraybackslash}m{.82\linewidth}}
$fHf$     & \begin{tikzpicture}[>=latex]
    \begin{axis}[
       width=0.4765\textwidth,
       height=0.25\textwidth,
  axis x line*=bottom,
  axis y line*=left,
  ylabel={Frequency} ,
  ymin = 0,
  xmin = -17,
  xmax = 3,
  xtick = {-16, -10, -4, 2},
  xticklabels={$10^{-16}$,$10^{-10}$,$10^{-4}$,$10^{2}$},
  xlabel=$\norm{\mat{H}-\mat{H}_{\te{est}}}_F/\norm{\mat{H}}_F$,
  ]
    \addplot[very thick,color=ourcolor] coordinates {
(-15.857254,1)
(-15.607254,4)
(-15.357254,42)
(-15.107254,118)
(-14.857254,318)
(-14.607254,522)
(-14.357254,693)
(-14.107254,869)
(-13.857254,846)
(-13.607254,795)
(-13.357254,768)
(-13.107254,755)
(-12.857254,629)
(-12.607254,610)
(-12.357254,517)
(-12.107254,442)
(-11.857254,422)
(-11.607254,327)
(-11.357254,258)
(-11.107254,222)
(-10.857254,196)
(-10.607254,123)
(-10.357254,107)
(-10.107254,90)
(-9.857254,75)
(-9.607254,46)
(-9.357254,42)
(-9.107254,33)
(-8.857254,38)
(-8.607254,19)
(-8.357254,16)
(-8.107254,13)
(-7.857254,16)
(-7.607254,7)
(-7.357254,4)
(-7.107254,6)
(-6.857254,4)
(-6.607254,2)
(-6.357254,4)
(-6.107254,1)
(-5.857254,0)
(-5.607254,0)
(-5.357254,0)
(-5.107254,0)
(-4.857254,0)
(-4.607254,0)
(-4.357254,0)
(-4.107254,0)
(-3.857254,0)
(-3.607254,0)
(-3.357254,0)
(-3.107254,0)
(-2.857254,0)
(-2.607254,0)
(-2.357254,0)
(-2.107254,0)
(-1.857254,0)
(-1.607254,0)
(-1.357254,0)
(-1.107254,0)
(-0.857254,0)
(-0.607254,0)
(-0.357254,0)
(-0.107254,0)
(0.142746,0)
(0.392746,0)
(0.642746,0)
(0.892746,0)
(1.142746,0)
(1.392746,0)
(1.642746,0)
(1.892746,0)
    };
    \addplot[very thick,color=dingcolor, dashed] coordinates {
(-15.857254,0)
(-15.607254,4)
(-15.357254,2)
(-15.107254,4)
(-14.857254,17)
(-14.607254,34)
(-14.357254,47)
(-14.107254,67)
(-13.857254,105)
(-13.607254,156)
(-13.357254,216)
(-13.107254,250)
(-12.857254,310)
(-12.607254,308)
(-12.357254,392)
(-12.107254,458)
(-11.857254,440)
(-11.607254,443)
(-11.357254,483)
(-11.107254,434)
(-10.857254,427)
(-10.607254,427)
(-10.357254,398)
(-10.107254,386)
(-9.857254,384)
(-9.607254,347)
(-9.357254,281)
(-9.107254,268)
(-8.857254,228)
(-8.607254,243)
(-8.357254,190)
(-8.107254,207)
(-7.857254,146)
(-7.607254,135)
(-7.357254,114)
(-7.107254,90)
(-6.857254,97)
(-6.607254,72)
(-6.357254,72)
(-6.107254,62)
(-5.857254,63)
(-5.607254,50)
(-5.357254,30)
(-5.107254,35)
(-4.857254,26)
(-4.607254,24)
(-4.357254,27)
(-4.107254,12)
(-3.857254,20)
(-3.607254,15)
(-3.357254,11)
(-3.107254,12)
(-2.857254,10)
(-2.607254,11)
(-2.357254,16)
(-2.107254,24)
(-1.857254,29)
(-1.607254,54)
(-1.357254,76)
(-1.107254,83)
(-0.857254,116)
(-0.607254,137)
(-0.357254,111)
(-0.107254,105)
(0.142746,80)
(0.392746,38)
(0.642746,26)
(0.892746,8)
(1.142746,4)
(1.392746,1)
(1.642746,1)
(1.892746,1)
};
    \end{axis}
    \begin{axis}[
       xshift=.5\textwidth,
       width=0.4765\textwidth,
       height=0.25\textwidth,
  axis x line*=bottom,
  axis y line*=left,
  ylabel={Frequency} ,
  ymin = 0,
  xmin = -17,
  xmax = 3,
  xtick = {-16, -10, -4, 2},
  xticklabels={$10^{-16}$,$10^{-10}$,$10^{-4}$,$10^{2}$},
  xlabel=$|f-f_{\te{est}}|/f$,
  ]
    \addplot[very thick,color=ourcolor] coordinates {
(-16.107254,0)
(-15.857254,113)
(-15.607254,132)
(-15.357254,286)
(-15.107254,471)
(-14.857254,720)
(-14.607254,853)
(-14.357254,853)
(-14.107254,859)
(-13.857254,857)
(-13.607254,722)
(-13.357254,626)
(-13.107254,553)
(-12.857254,462)
(-12.607254,418)
(-12.357254,374)
(-12.107254,299)
(-11.857254,275)
(-11.607254,202)
(-11.357254,180)
(-11.107254,151)
(-10.857254,118)
(-10.607254,85)
(-10.357254,67)
(-10.107254,51)
(-9.857254,41)
(-9.607254,33)
(-9.357254,27)
(-9.107254,16)
(-8.857254,23)
(-8.607254,14)
(-8.357254,10)
(-8.107254,4)
(-7.857254,4)
(-7.607254,3)
(-7.357254,3)
(-7.107254,2)
(-6.857254,2)
(-6.607254,3)
(-6.357254,0)
(-6.107254,1)
(-5.857254,0)
(-5.607254,0)
(-5.357254,0)
(-5.107254,0)
(-4.857254,0)
(-4.607254,0)
(-4.357254,0)
(-4.107254,0)
(-3.857254,0)
(-3.607254,0)
(-3.357254,0)
(-3.107254,0)
(-2.857254,0)
(-2.607254,0)
(-2.357254,0)
(-2.107254,0)
(-1.857254,0)
(-1.607254,0)
(-1.357254,0)
(-1.107254,0)
(-0.857254,0)
(-0.607254,0)
(-0.357254,0)
(-0.107254,0)
(0.142746,0)
(0.392746,0)
(0.642746,0)
(0.892746,0)
(1.142746,0)
(1.392746,0)
(1.642746,0)
(1.892746,0)
    };
    \addplot[very thick,color=dingcolor, dashed] coordinates {
(-16.107254,0)
(-15.857254,11)
(-15.607254,21)
(-15.357254,28)
(-15.107254,68)
(-14.857254,101)
(-14.607254,171)
(-14.357254,260)
(-14.107254,334)
(-13.857254,442)
(-13.607254,521)
(-13.357254,567)
(-13.107254,624)
(-12.857254,612)
(-12.607254,619)
(-12.357254,615)
(-12.107254,580)
(-11.857254,530)
(-11.607254,463)
(-11.357254,402)
(-11.107254,345)
(-10.857254,321)
(-10.607254,234)
(-10.357254,223)
(-10.107254,172)
(-9.857254,116)
(-9.607254,140)
(-9.357254,81)
(-9.107254,71)
(-8.857254,69)
(-8.607254,47)
(-8.357254,44)
(-8.107254,44)
(-7.857254,37)
(-7.607254,26)
(-7.357254,18)
(-7.107254,25)
(-6.857254,20)
(-6.607254,13)
(-6.357254,14)
(-6.107254,12)
(-5.857254,8)
(-5.607254,8)
(-5.357254,7)
(-5.107254,6)
(-4.857254,5)
(-4.607254,5)
(-4.357254,5)
(-4.107254,5)
(-3.857254,1)
(-3.607254,0)
(-3.357254,2)
(-3.107254,4)
(-2.857254,3)
(-2.607254,2)
(-2.357254,1)
(-2.107254,4)
(-1.857254,7)
(-1.607254,5)
(-1.357254,13)
(-1.107254,20)
(-0.857254,43)
(-0.607254,45)
(-0.357254,85)
(-0.107254,194)
(0.142746,169)
(0.392746,184)
(0.642746,47)
(0.892746,29)
(1.142746,19)
(1.392746,7)
(1.642746,9)
(1.892746,2)
};
\end{axis}
\end{tikzpicture}\\%
$Hf$      & \begin{tikzpicture}[>=latex]
    \begin{axis}[
       width=0.4765\textwidth,
       height=0.25\textwidth,
  axis x line*=bottom,
  axis y line*=left,
  ylabel={Frequency} ,
  ymin = 0,
  xmin = -17,
  xmax = 3,
  xtick = {-16, -10, -4, 2},
  xticklabels={$10^{-16}$,$10^{-10}$,$10^{-4}$,$10^{2}$},
  xlabel=$\norm{\mat{H}-\mat{H}_{\te{est}}}_F/\norm{\mat{H}}_F$,
  ]
    \addplot[very thick,color=ourcolor] coordinates {
(-16.357254,0)
(-16.107254,4)
(-15.857254,8)
(-15.607254,83)
(-15.357254,246)
(-15.107254,645)
(-14.857254,1086)
(-14.607254,1428)
(-14.357254,1404)
(-14.107254,1207)
(-13.857254,933)
(-13.607254,765)
(-13.357254,509)
(-13.107254,472)
(-12.857254,332)
(-12.607254,252)
(-12.357254,162)
(-12.107254,117)
(-11.857254,95)
(-11.607254,70)
(-11.357254,40)
(-11.107254,35)
(-10.857254,28)
(-10.607254,19)
(-10.357254,18)
(-10.107254,12)
(-9.857254,8)
(-9.607254,4)
(-9.357254,2)
(-9.107254,6)
(-8.857254,1)
(-8.607254,2)
(-8.357254,0)
(-8.107254,0)
(-7.857254,2)
(-7.607254,0)
(-7.357254,1)
(-7.107254,2)
(-6.857254,1)
(-6.607254,1)
(-6.357254,0)
(-6.107254,0)
(-5.857254,0)
(-5.607254,0)
(-5.357254,0)
(-5.107254,0)
(-4.857254,0)
(-4.607254,0)
(-4.357254,0)
(-4.107254,0)
(-3.857254,0)
(-3.607254,0)
(-3.357254,0)
(-3.107254,0)
(-2.857254,0)
(-2.607254,0)
(-2.357254,0)
(-2.107254,0)
(-1.857254,0)
(-1.607254,0)
(-1.357254,0)
(-1.107254,0)
(-0.857254,0)
(-0.607254,0)
(-0.357254,0)
(-0.107254,0)
(0.142746,0)
(0.392746,0)
(0.642746,0)
(0.892746,0)
(1.142746,0)
(1.392746,0)
(1.642746,0)
(1.892746,0)
    };
    \addplot[very thick,color=dingcolor, dashed] coordinates {
(-16.357254,0)
(-16.107254,0)
(-15.857254,3)
(-15.607254,15)
(-15.357254,64)
(-15.107254,163)
(-14.857254,270)
(-14.607254,415)
(-14.357254,548)
(-14.107254,676)
(-13.857254,749)
(-13.607254,820)
(-13.357254,808)
(-13.107254,782)
(-12.857254,764)
(-12.607254,687)
(-12.357254,505)
(-12.107254,516)
(-11.857254,400)
(-11.607254,348)
(-11.357254,272)
(-11.107254,202)
(-10.857254,177)
(-10.607254,166)
(-10.357254,130)
(-10.107254,86)
(-9.857254,77)
(-9.607254,64)
(-9.357254,43)
(-9.107254,45)
(-8.857254,38)
(-8.607254,24)
(-8.357254,17)
(-8.107254,19)
(-7.857254,14)
(-7.607254,14)
(-7.357254,14)
(-7.107254,6)
(-6.857254,11)
(-6.607254,7)
(-6.357254,6)
(-6.107254,3)
(-5.857254,7)
(-5.607254,5)
(-5.357254,0)
(-5.107254,4)
(-4.857254,1)
(-4.607254,0)
(-4.357254,4)
(-4.107254,1)
(-3.857254,1)
(-3.607254,1)
(-3.357254,0)
(-3.107254,1)
(-2.857254,2)
(-2.607254,1)
(-2.357254,0)
(-2.107254,1)
(-1.857254,0)
(-1.607254,2)
(-1.357254,0)
(-1.107254,0)
(-0.857254,0)
(-0.607254,0)
(-0.357254,0)
(-0.107254,0)
(0.142746,1)
(0.392746,0)
(0.642746,0)
(0.892746,0)
(1.142746,0)
(1.392746,0)
(1.642746,0)
(1.892746,0)
};
    \end{axis}
    \begin{axis}[
       xshift=.5\textwidth,
       width=0.4765\textwidth,
       height=0.25\textwidth,
  axis x line*=bottom,
  axis y line*=left,
  ylabel={Frequency} ,
  ymin = 0,
  xmin = -17,
  xmax = 3,
  xtick = {-16, -10, -4, 2},
  xticklabels={$10^{-16}$,$10^{-10}$,$10^{-4}$,$10^{2}$},
  xlabel=$|f-f_{\te{est}}|/f$,
  ]
    \addplot[very thick,color=ourcolor] coordinates {
(-16.357254,0)
(-16.107254,0)
(-15.857254,217)
(-15.607254,348)
(-15.357254,592)
(-15.107254,1012)
(-14.857254,1378)
(-14.607254,1492)
(-14.357254,1336)
(-14.107254,1024)
(-13.857254,807)
(-13.607254,544)
(-13.357254,363)
(-13.107254,264)
(-12.857254,165)
(-12.607254,98)
(-12.357254,55)
(-12.107254,42)
(-11.857254,27)
(-11.607254,15)
(-11.357254,19)
(-11.107254,9)
(-10.857254,4)
(-10.607254,2)
(-10.357254,4)
(-10.107254,1)
(-9.857254,1)
(-9.607254,2)
(-9.357254,0)
(-9.107254,0)
(-8.857254,0)
(-8.607254,0)
(-8.357254,0)
(-8.107254,0)
(-7.857254,0)
(-7.607254,0)
(-7.357254,0)
(-7.107254,0)
(-6.857254,0)
(-6.607254,0)
(-6.357254,0)
(-6.107254,0)
(-5.857254,0)
(-5.607254,0)
(-5.357254,0)
(-5.107254,0)
(-4.857254,0)
(-4.607254,0)
(-4.357254,0)
(-4.107254,0)
(-3.857254,0)
(-3.607254,0)
(-3.357254,0)
(-3.107254,0)
(-2.857254,0)
(-2.607254,0)
(-2.357254,0)
(-2.107254,0)
(-1.857254,0)
(-1.607254,0)
(-1.357254,0)
(-1.107254,0)
(-0.857254,0)
(-0.607254,0)
(-0.357254,0)
(-0.107254,0)
(0.142746,0)
(0.392746,0)
(0.642746,0)
(0.892746,0)
(1.142746,0)
(1.392746,0)
(1.642746,0)
(1.892746,0)
    };
    \addplot[very thick,color=dingcolor, dashed] coordinates {
(-16.357254,0)
(-16.107254,0)
(-15.857254,35)
(-15.607254,56)
(-15.357254,93)
(-15.107254,167)
(-14.857254,245)
(-14.607254,415)
(-14.357254,634)
(-14.107254,771)
(-13.857254,911)
(-13.607254,877)
(-13.357254,871)
(-13.107254,777)
(-12.857254,696)
(-12.607254,595)
(-12.357254,511)
(-12.107254,446)
(-11.857254,369)
(-11.607254,294)
(-11.357254,208)
(-11.107254,171)
(-10.857254,126)
(-10.607254,143)
(-10.357254,91)
(-10.107254,83)
(-9.857254,83)
(-9.607254,43)
(-9.357254,28)
(-9.107254,38)
(-8.857254,29)
(-8.607254,14)
(-8.357254,25)
(-8.107254,14)
(-7.857254,20)
(-7.607254,5)
(-7.357254,11)
(-7.107254,14)
(-6.857254,6)
(-6.607254,8)
(-6.357254,3)
(-6.107254,5)
(-5.857254,3)
(-5.607254,2)
(-5.357254,4)
(-5.107254,3)
(-4.857254,2)
(-4.607254,0)
(-4.357254,2)
(-4.107254,1)
(-3.857254,2)
(-3.607254,0)
(-3.357254,1)
(-3.107254,0)
(-2.857254,1)
(-2.607254,1)
(-2.357254,0)
(-2.107254,1)
(-1.857254,1)
(-1.607254,0)
(-1.357254,0)
(-1.107254,0)
(-0.857254,0)
(-0.607254,0)
(-0.357254,1)
(-0.107254,2)
(0.142746,1)
(0.392746,0)
(0.642746,0)
(0.892746,0)
(1.142746,0)
(1.392746,0)
(1.642746,0)
(1.892746,0)
};
\end{axis}
\end{tikzpicture}\\%
$f_1Hf_2$ & \begin{tikzpicture}[>=latex]
    \begin{axis}[
       width=0.4765\textwidth,
       height=0.25\textwidth,
  axis x line*=bottom,
  axis y line*=left,
  ylabel={Frequency} ,
  ymin = 0,
  xmin = -17,
  xmax = 3,
  xtick = {-16, -10, -4, 2},
  xticklabels={$10^{-16}$,$10^{-10}$,$10^{-4}$,$10^{2}$},
  xlabel=$\norm{\mat{H}-\mat{H}_{\te{est}}}_F/\norm{\mat{H}}_F$,
  ]
    \addplot[very thick,color=ourcolor] coordinates {
(-16.857254,0)
(-16.607254,0)
(-16.357254,1)
(-16.107254,2)
(-15.857254,4)
(-15.607254,23)
(-15.357254,100)
(-15.107254,297)
(-14.857254,489)
(-14.607254,733)
(-14.357254,925)
(-14.107254,1019)
(-13.857254,939)
(-13.607254,910)
(-13.357254,809)
(-13.107254,674)
(-12.857254,558)
(-12.607254,462)
(-12.357254,395)
(-12.107254,302)
(-11.857254,243)
(-11.607254,187)
(-11.357254,165)
(-11.107254,133)
(-10.857254,110)
(-10.607254,83)
(-10.357254,67)
(-10.107254,70)
(-9.857254,54)
(-9.607254,52)
(-9.357254,31)
(-9.107254,26)
(-8.857254,19)
(-8.607254,20)
(-8.357254,15)
(-8.107254,8)
(-7.857254,12)
(-7.607254,14)
(-7.357254,10)
(-7.107254,6)
(-6.857254,5)
(-6.607254,5)
(-6.357254,1)
(-6.107254,4)
(-5.857254,4)
(-5.607254,1)
(-5.357254,3)
(-5.107254,5)
(-4.857254,0)
(-4.607254,0)
(-4.357254,0)
(-4.107254,0)
(-3.857254,0)
(-3.607254,2)
(-3.357254,0)
(-3.107254,0)
(-2.857254,1)
(-2.607254,0)
(-2.357254,0)
(-2.107254,1)
(-1.857254,0)
(-1.607254,0)
(-1.357254,0)
(-1.107254,0)
(-0.857254,0)
(-0.607254,0)
(-0.357254,0)
(-0.107254,0)
(0.142746,0)
(0.392746,0)
(0.642746,0)
(0.892746,0)
(1.142746,0)
(1.392746,0)
(1.642746,0)
(1.892746,0)
    };
    \addplot[very thick,color=dingcolor, dashed] coordinates {
(-16.857254,0)
(-16.607254,0)
(-16.357254,0)
(-16.107254,1)
(-15.857254,32)
(-15.607254,130)
(-15.357254,440)
(-15.107254,804)
(-14.857254,996)
(-14.607254,1042)
(-14.357254,1063)
(-14.107254,1039)
(-13.857254,970)
(-13.607254,814)
(-13.357254,683)
(-13.107254,502)
(-12.857254,386)
(-12.607254,320)
(-12.357254,218)
(-12.107254,176)
(-11.857254,113)
(-11.607254,104)
(-11.357254,53)
(-11.107254,29)
(-10.857254,19)
(-10.607254,25)
(-10.357254,9)
(-10.107254,12)
(-9.857254,5)
(-9.607254,5)
(-9.357254,4)
(-9.107254,2)
(-8.857254,3)
(-8.607254,0)
(-8.357254,1)
(-8.107254,0)
(-7.857254,0)
(-7.607254,0)
(-7.357254,0)
(-7.107254,0)
(-6.857254,0)
(-6.607254,0)
(-6.357254,0)
(-6.107254,0)
(-5.857254,0)
(-5.607254,0)
(-5.357254,0)
(-5.107254,0)
(-4.857254,0)
(-4.607254,0)
(-4.357254,0)
(-4.107254,0)
(-3.857254,0)
(-3.607254,0)
(-3.357254,0)
(-3.107254,0)
(-2.857254,0)
(-2.607254,0)
(-2.357254,0)
(-2.107254,0)
(-1.857254,0)
(-1.607254,0)
(-1.357254,0)
(-1.107254,0)
(-0.857254,0)
(-0.607254,0)
(-0.357254,0)
(-0.107254,0)
(0.142746,0)
(0.392746,0)
(0.642746,0)
(0.892746,0)
(1.142746,0)
(1.392746,0)
(1.642746,0)
(1.892746,0)
};
    \end{axis}
    \begin{axis}[
       xshift=.5\textwidth,
       width=0.4765\textwidth,
       height=0.25\textwidth,
  axis x line*=bottom,
  axis y line*=left,
  ylabel={Frequency} ,
  ymin = 0,
  xmin = -17,
  xmax = 3,
  xtick = {-16, -10, -4, 2},
  xticklabels={$10^{-16}$,$10^{-10}$,$10^{-4}$,$10^{2}$},
  xlabel=$|f-f_{\te{est}}|/f$,
  ]
    \addplot[very thick,color=ourcolor] coordinates {
(-16.857254,0)
(-16.607254,0)
(-16.357254,0)
(-16.107254,6)
(-15.857254,12)
(-15.607254,29)
(-15.357254,87)
(-15.107254,245)
(-14.857254,537)
(-14.607254,746)
(-14.357254,992)
(-14.107254,1034)
(-13.857254,1013)
(-13.607254,978)
(-13.357254,751)
(-13.107254,654)
(-12.857254,532)
(-12.607254,431)
(-12.357254,380)
(-12.107254,275)
(-11.857254,199)
(-11.607254,186)
(-11.357254,155)
(-11.107254,134)
(-10.857254,99)
(-10.607254,78)
(-10.357254,67)
(-10.107254,61)
(-9.857254,48)
(-9.607254,40)
(-9.357254,39)
(-9.107254,41)
(-8.857254,23)
(-8.607254,16)
(-8.357254,19)
(-8.107254,18)
(-7.857254,12)
(-7.607254,11)
(-7.357254,5)
(-7.107254,9)
(-6.857254,8)
(-6.607254,6)
(-6.357254,3)
(-6.107254,1)
(-5.857254,4)
(-5.607254,3)
(-5.357254,4)
(-5.107254,0)
(-4.857254,1)
(-4.607254,1)
(-4.357254,0)
(-4.107254,2)
(-3.857254,1)
(-3.607254,0)
(-3.357254,0)
(-3.107254,0)
(-2.857254,1)
(-2.607254,0)
(-2.357254,0)
(-2.107254,0)
(-1.857254,0)
(-1.607254,0)
(-1.357254,0)
(-1.107254,0)
(-0.857254,1)
(-0.607254,0)
(-0.357254,0)
(-0.107254,0)
(0.142746,0)
(0.392746,0)
(0.642746,0)
(0.892746,0)
(1.142746,0)
(1.392746,0)
(1.642746,0)
(1.892746,0)
    };
    \addplot[very thick,color=dingcolor, dashed] coordinates {
(-16.857254,0)
(-16.607254,0)
(-16.357254,0)
(-16.107254,0)
(-15.857254,1)
(-15.607254,12)
(-15.357254,20)
(-15.107254,42)
(-14.857254,130)
(-14.607254,298)
(-14.357254,484)
(-14.107254,657)
(-13.857254,836)
(-13.607254,809)
(-13.357254,829)
(-13.107254,788)
(-12.857254,690)
(-12.607254,610)
(-12.357254,572)
(-12.107254,481)
(-11.857254,411)
(-11.607254,335)
(-11.357254,298)
(-11.107254,217)
(-10.857254,226)
(-10.607254,179)
(-10.357254,132)
(-10.107254,130)
(-9.857254,124)
(-9.607254,100)
(-9.357254,78)
(-9.107254,81)
(-8.857254,54)
(-8.607254,60)
(-8.357254,58)
(-8.107254,40)
(-7.857254,35)
(-7.607254,23)
(-7.357254,15)
(-7.107254,17)
(-6.857254,19)
(-6.607254,20)
(-6.357254,8)
(-6.107254,11)
(-5.857254,12)
(-5.607254,9)
(-5.357254,3)
(-5.107254,4)
(-4.857254,5)
(-4.607254,4)
(-4.357254,3)
(-4.107254,2)
(-3.857254,1)
(-3.607254,4)
(-3.357254,4)
(-3.107254,1)
(-2.857254,2)
(-2.607254,1)
(-2.357254,2)
(-2.107254,1)
(-1.857254,1)
(-1.607254,2)
(-1.357254,1)
(-1.107254,1)
(-0.857254,2)
(-0.607254,1)
(-0.357254,1)
(-0.107254,1)
(0.142746,0)
(0.392746,0)
(0.642746,1)
(0.892746,0)
(1.142746,0)
(1.392746,0)
(1.642746,0)
(1.892746,0)
};
\end{axis}
\end{tikzpicture}\\%
          & \includestandalone[width=0.3\textwidth]{graphs/histogram/histogram_legend}\
\end{tabular}
\caption{Error histogram for 10,000 randomly generated problem instances for the partially
calibrated cases. Top to bottom: $fHf$, $Hf$, $f_1Hf_2$.}
\vspace{-3mm}
\label{fig:exp1}
\end{figure*}

\section{Incorporating the IMU data}
We follow the approach used in~\cite{ding-etal-2019-iccv}.
A general homography~$\mat{H}$ has eight degrees of freedom, fulfilling the Direct Linear
Transform (DLT) constraint,
\begin{equation}\label{eq:dlt}
    \vec{x}_2 \sim \mat{H}\vec{x}_1,
\end{equation}
for two point correspondences~$\vec{x}_1\leftrightarrow\vec{x}_2$ on a common scene plane.
The relation~\eqref{eq:dlt} yields two linearly independent equations, hence a minimum of
four point correspondences are necessary in order to estimate the homography. In the calibrated
case, with intrinsic parameters encoded in the calibration matrices~$\mat{K}_1$ and~$\mat{K}_2$,
respectively, one obtains
\begin{equation}
    \Hc \sim \mat{K}_2\mat{H}\mat{K}_1^{-1},
\end{equation}
where the Euclidean homography~$\Hc$ can be written as
\begin{equation}
    \Hc \sim \mat{R} + \fr{1}{d}\vec{t}\vec{n}^\T,
\end{equation}
where $\mat{R}$ is the relative rotation between the views,~$\vec{n}$ is the plane normal,
and~$\vec{t}$ is the relative translation. The depth parameter~$d$ is the distance from
the first camera center to the scene plane.

\begin{figure*}[t!]
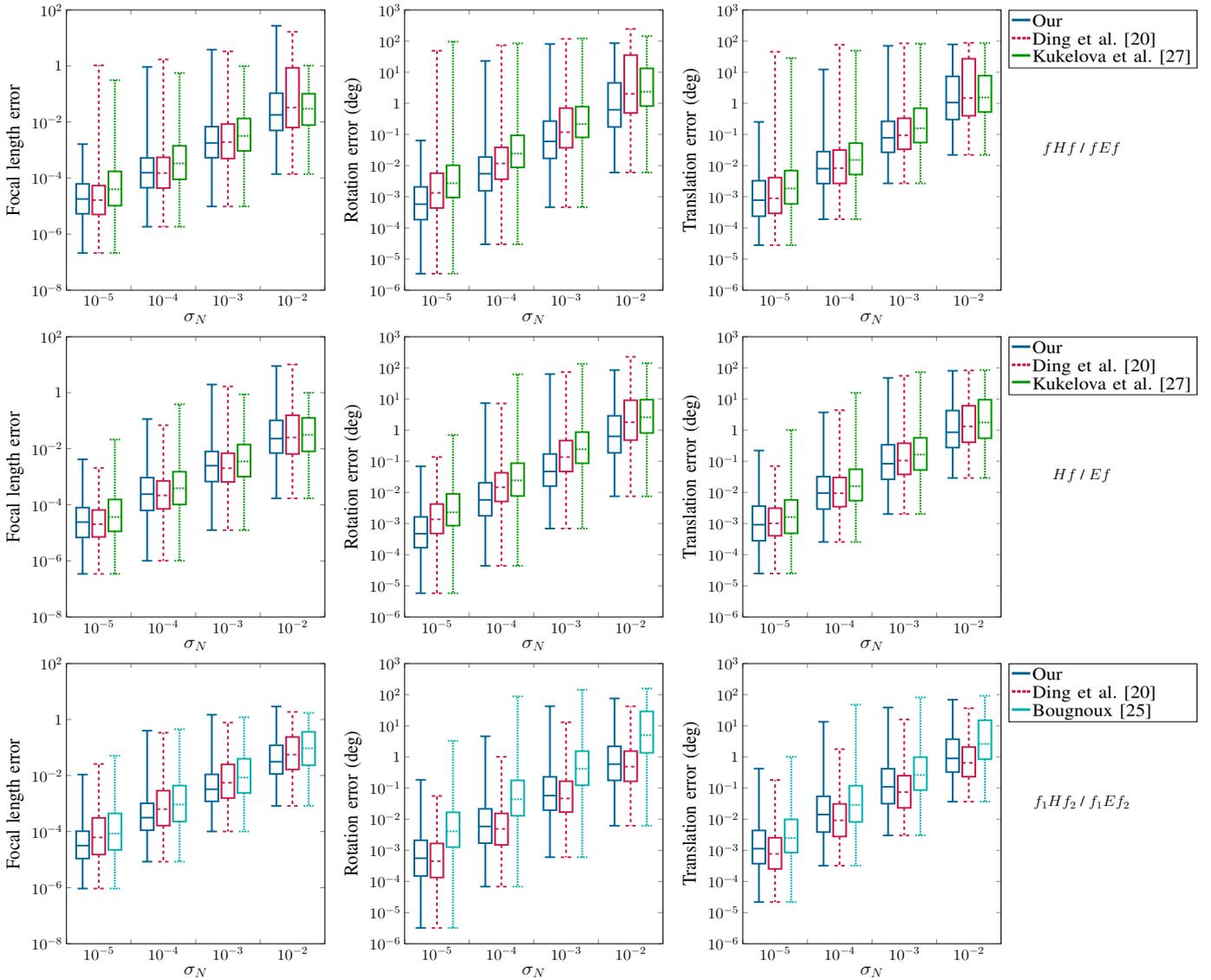

\includestandalone[height=5cm,width=\linewidth]{graphs/box/box_fHf}\\%
\includestandalone[height=5cm,width=\linewidth]{graphs/box/box_Hf}\\%
\includestandalone[height=5cm,width=\linewidth]{graphs/box/box_f1Hf2}%
\caption{Noise sensitivity comparison for Gaussian noise with standard deviation~$\sigma_N$.
A total of 1,000 random problem instances were generated per noise level.}
\label{fig:noise}
\end{figure*}

We consider the case of a common reference direction, which eliminates two degrees of freedom,
see \cref{fig:1}. Therefore, after a suitable change of coordinates, we may assume that
\begin{equation}\label{eq:Hy}
    \Hy \sim \mat{R}_y + \fr{1}{d'}\vec{t}'\vec{n}^{'\T}\;.
\end{equation}
Hence, the DLT equations~\eqref{eq:dlt} can be written as
\begin{equation}\label{eq:calib}
    \mat{R}_2^\T\mat{K}_2^{-1}\vec{x}_2\sim \Hy\mat{R}_1^\T\mat{K}_1^{-1}\vec{x}_1,
\end{equation}
where
$\vec{R}_y$ is a rotation about the $y$-axis (gravitational direction). The known matrices~$\mat{R}_1$ and
$\mat{R}_2$ are given by the IMU data, and consists of the pitch and roll angles for
the first and second camera position.

The relation between the original homography~$\mat{H}$, from~\eqref{eq:dlt}, and $\Hy$,
from~\eqref{eq:Hy}, is thus given by
\begin{equation}\label{eq:HvsHy}
    \Hy\sim\mat{R}_2^\T\mat{K}^{-1}_2\mat{H}\mat{K}_1\mat{R}_1\;.
\end{equation}
Furthermore, the relative rotation~$\mat{R}$ and the relative translation~$\mat{t}$,
are given by
\begin{equation}
    \mat{R} = \mat{R}_2\mat{R}_y\mat{R}_1^\T
    \quad\te{and}\quad
    \fr{\mat{t}}{d} = \mat{R}_2\fr{\mat{t}'}{d'},
\end{equation}
with the modified plane normal~$\mat{n}'=\mat{R}_1\mat{n}$.

\subsection{Navigation using the ground plane}
By introducing the auxiliary variables
\begin{equation}\label{eq:yi}
    \vec{y}_i = \mat{R}_j^\T\mat{K}_j^{-1}\vec{x}_i,
\end{equation}
one may reduce~\eqref{eq:calib}
to~$\vec{y}_2\sim\Hy\vec{y}_1$. For an arbitrary plane normal~$\mat{n}'$,
the matrix~$\Hy$, defined as in~\eqref{eq:Hy}, has~6~degrees of freedom (DoF).
However, if we constrain ourselves to navigating using the ground plane, then the plane normal
is uniquely defined (up to scale), with $\mat{n}'=(0,\,1,\,0)^\T$. Note, that
we can only recover the translation up to scale, hence we may assume that the depth~$d'=1$.
Parameterizing the rotation matrix as
\begin{equation}
    \mat{R}_y =
    \begin{bmatrix}
        \cos{\theta} & 0 & \sin{\theta}\\
        0            & 1 & 0 \\
       -\sin{\theta} & 0 & \cos{\theta} \\
    \end{bmatrix},
\end{equation}
we may write
\begin{equation}\label{eq:Hyparam}
    \Hy =
    \begin{bmatrix}
        h_1 & h_3 & h_2 \\
        0   & h_4 & 0 \\
       -h_2 & h_5 & h_1
    \end{bmatrix},
\end{equation}
where~$\mat{R}$ and~$\mat{t}$ can be extracted directly through the entries~$h_i$,
given by
\begin{equation}\label{eq:Ry}
    \mat{R} =
    \begin{bmatrix}
        h_1 & 0 & h_2 \\
        0   & 1 & 0 \\
       -h_2 & 0 & h_1
    \end{bmatrix}
    \te{ and }
    \mat{t} =
    \begin{bmatrix}
        h_3 \\
        h_4-1 \\
        h_5
    \end{bmatrix},
\end{equation}
where the trigonometric constraint~$h_1^2+h_2^2=1$ must be enforced to get a
valid rotation matrix.

\subsection{The calibrated case}
In~\cite{saurer-etal-2017}, the authors construct a minimal solver for the calibrated case,
with 4 DoF (three translation components, and the unknown rotation about the ground floor normal).
As a consequence, one only needs two point correspondences, which give rise to four linearly
independent equations. By parameterizing~$\Hy$ as in~\eqref{eq:Hy}, it is possible to form the
linear system\footnote{In~\cite{saurer-etal-2017} the camera is aligned with the $z$-axis, but the
same procedure can be repeated using the~$y$-axis instead.}
\begin{equation}
    \mat{A}\vec{h} = \vec{0},
\end{equation}
where $\mat{A}$ is a $4\times 5$ matrix and~$\vec{h}$ contains the~$h_i$. For non-degenerate configurations, the matrix~$\mat{A}$
has a one-dimensional nullspace, which can be obtained using singular value decomposition~(SVD).
One may fix the scale of~$\Hy$, in order to acquire valid rotation parameters. This is achieved by
using the trigonometric constraint introduced through the parameterization of the rotation
matrix, $h_1^2+h_2^2=1$, as in~\eqref{eq:Ry}. This approach leaves two possible
solutions~\cite{saurer-etal-2017}.

\vspace*{0.2em}
\section{Partially calibrated cases for ground plane navigation}
We will extend the ground plane solver to three partially calibrated cases, also considered
in~\cite{ding-etal-2019-iccv}.

\subsection{Equal and unknown focal length (\texorpdfstring{$fHf$}{fHf}, 2.5-point)}
Let us parameterize the inverse of the unknown calibration matrix
as~$\mat{K}^{-1} = \diag(1,\,1,\,w)$, and consider the
rectified points~\eqref{eq:yi}, \pagebreak which now depend linearly on the unknown parameter~$w$.
Parameterizing~$\Hy$ as in~\eqref{eq:Hyparam}, it is clear that
the equations obtained from~\eqref{eq:calib} are quadratic in $h_i$, $i=1,\ldots,5$ and~$w$.
Using 2.5-point correspondences,~\ie{} using 3-point correspondences and discarding one of
the DLT equations, together with the constraint $h_1^2+h_2^2=1$, gives  six equations and six
unknowns.

This system of equations has infinitely many solutions, if we allow~\mbox{$w=0$}. Such solutions, however,
do not yield geometrically meaningful reconstructions, and should therefore be excluded.
This can be achieved using saturation, through the method suggested in~\cite{larsson2017ICCV}.

To generate a polynomial solver, we use the automatic generator proposed in~\cite{larsson2016eccv}, from which we have 14 solutions, and 
an elimination template of size~$10\times 17$.
Exploiting symmetry, the solutions can be obtained by solving a $7\times 7$ eigenvalue problem.

\subsection{One unknown focal length (\texorpdfstring{$Hf$}{HF}, 2.5-point)}
We may use the same approach as in the previous case, but may
pre-compute~$\mat{y}_1=\mat{R}_1^\T\mat{K}_1^{-1}\vec{x}_i$, to reduce the number of unknown
coefficients in the elimination template.
Due to this approach, we are able to reduce the eigenvalue problem to a $4\times 4$
matrix\footnote{In fact, one may use the quartic formula for root finding instead, to gain a bit of extra performance.}, despite having the same amount of solutions and elimination template
initially.

\subsection{Different and unknown focal lengths (\texorpdfstring{$f_1Hf_2$}{f1Hf2}, 3-point)}
Using the same approach as previously for this case resulted in a large elimination template,
and in order to reduce the template size we opted for a different approach. Similar
to~\cite{ding-etal-2019-iccv} we parameterize the nullspace of the homography~\eqref{eq:dlt}.
In general, three point correspondences yield six linearly independent equations, hence
the corresponding nullspace is of dimension~3 and can be parameterized using
\begin{equation}
    \mat{H} = \alpha_0\mat{H}_0 + \alpha_1\mat{H}_1 + \alpha_2\mat{H}_2.
\end{equation}
We may fix the scale by letting~$\alpha_0=1$. As in the previous case, we may parameterize
the calibration matrices $\mat{K}_1 = \diag(1,\,1,\,w_1)$
and~$\mat{K}_2^{-1} = \diag(1,\,1,\,w_2)$. Inserting this into~\eqref{eq:HvsHy} yields
a parameterization of~$\Hy$ in the unknowns $\alpha_1,\,\alpha_2,\,w_1,\,w_2$ of cubic degree.

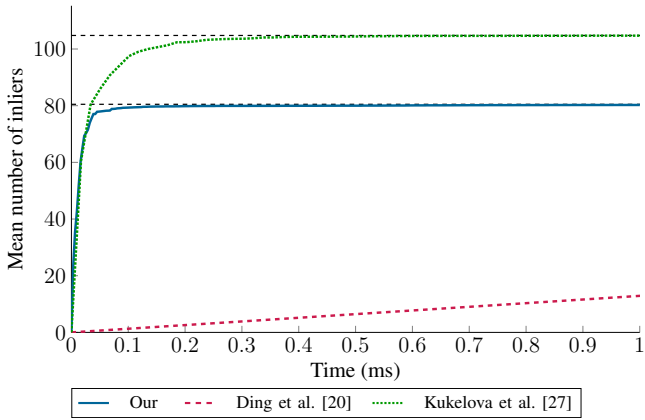
\begin{figure}[t]
\centering
\begin{tikzpicture}[>=latex]
    \begin{axis}[
       width=0.8\textwidth,
       height=0.5\textwidth,
  axis x line*=bottom,
  axis y line*=left,
  ylabel={Mean number of inliers} ,
  ymin = 0,
  xmin = 0,
  xmax = 1,
  label style={
    font=\Large
  },
  tick label style={
    font=\Large,
  },
  xlabel=Time (ms),
  ]

    \addplot[\mythicknesssmall,color=smallcolor1, dashed] coordinates {
(0.000000,8.049000e+01)
(20.000000,8.049000e+01)
};
    \addplot[\mythicknesssmall,color=smallcolor2, dashed] coordinates {
(0.000000,1.047800e+02)
(20.000000,1.047800e+02)
};

    \addplot[\mythickness,color=ourcolor] coordinates {
(0.000000,0)
(0.003227,2.361000e+01)
(0.006453,3.553000e+01)
(0.009680,4.275000e+01)
(0.012906,5.194000e+01)
(0.016133,6.023000e+01)
(0.019359,65)
(0.022586,6.939000e+01)
(0.025813,7.030000e+01)
(0.029039,7.154000e+01)
(0.032266,7.395000e+01)
(0.035492,7.554000e+01)
(0.038719,7.705000e+01)
(0.041945,7.707000e+01)
(0.045172,7.785000e+01)
(0.048399,7.790000e+01)
(0.051625,7.798000e+01)
(0.054852,7.806000e+01)
(0.058078,7.814000e+01)
(0.061305,7.822000e+01)
(0.064531,7.824000e+01)
(0.067758,7.828000e+01)
(0.070985,7.885000e+01)
(0.074211,7.886000e+01)
(0.077438,7.892000e+01)
(0.080664,7.906000e+01)
(0.083891,7.910000e+01)
(0.087117,7.914000e+01)
(0.090344,7.925000e+01)
(0.093571,7.927000e+01)
(0.096797,7.930000e+01)
(0.100024,7.932000e+01)
(0.103250,7.936000e+01)
(0.106477,7.938000e+01)
(0.109703,7.942000e+01)
(0.112930,7.944000e+01)
(0.116157,7.946000e+01)
(0.119383,7.949000e+01)
(0.122610,7.950000e+01)
(0.125836,7.959000e+01)
(0.129063,7.961000e+01)
(0.132289,7.961000e+01)
(0.135516,7.965000e+01)
(0.138743,7.967000e+01)
(0.141969,7.968000e+01)
(0.145196,7.969000e+01)
(0.148422,7.971000e+01)
(0.151649,7.971000e+01)
(0.154875,7.971000e+01)
(0.158102,7.972000e+01)
(0.161329,7.972000e+01)
(0.164555,7.975000e+01)
(0.167782,7.976000e+01)
(0.171008,7.976000e+01)
(0.174235,7.976000e+01)
(0.177461,7.976000e+01)
(0.180688,7.976000e+01)
(0.183915,7.978000e+01)
(0.187141,7.978000e+01)
(0.190368,7.979000e+01)
(0.193594,7.981000e+01)
(0.196821,7.982000e+01)
(0.200047,7.983000e+01)
(0.203274,7.983000e+01)
(0.206501,7.984000e+01)
(0.209727,7.986000e+01)
(0.212954,7.986000e+01)
(0.216180,7.987000e+01)
(0.219407,7.987000e+01)
(0.222633,7.987000e+01)
(0.225860,7.988000e+01)
(0.229087,7.988000e+01)
(0.232313,7.989000e+01)
(0.235540,7.989000e+01)
(0.238766,7.989000e+01)
(0.241993,7.989000e+01)
(0.245219,7.989000e+01)
(0.248446,7.989000e+01)
(0.251673,7.989000e+01)
(0.254899,7.992000e+01)
(0.258126,7.992000e+01)
(0.261352,7.992000e+01)
(0.264579,7.992000e+01)
(0.267805,7.993000e+01)
(0.271032,7.993000e+01)
(0.274259,7.993000e+01)
(0.277485,7.993000e+01)
(0.280712,7.993000e+01)
(0.283938,7.993000e+01)
(0.287165,7.993000e+01)
(0.290391,7.993000e+01)
(0.293618,7.993000e+01)
(0.296845,7.993000e+01)
(0.300071,7.993000e+01)
(0.303298,7.993000e+01)
(0.306524,7.993000e+01)
(0.309751,7.993000e+01)
(0.312977,7.993000e+01)
(0.316204,7.994000e+01)
(0.319431,7.994000e+01)
(0.322657,7.994000e+01)
(0.325884,7.994000e+01)
(0.329110,7.994000e+01)
(0.332337,7.994000e+01)
(0.335563,7.994000e+01)
(0.338790,7.994000e+01)
(0.342017,7.994000e+01)
(0.345243,7.994000e+01)
(0.348470,7.994000e+01)
(0.351696,7.994000e+01)
(0.354923,7.994000e+01)
(0.358149,7.994000e+01)
(0.361376,7.994000e+01)
(0.364603,7.994000e+01)
(0.367829,7.995000e+01)
(0.371056,7.996000e+01)
(0.374282,7.996000e+01)
(0.377509,7.996000e+01)
(0.380735,7.996000e+01)
(0.383962,7.996000e+01)
(0.387189,7.996000e+01)
(0.390415,7.996000e+01)
(0.393642,7.996000e+01)
(0.396868,7.996000e+01)
(0.400095,7.996000e+01)
(0.403322,7.997000e+01)
(0.406548,7.997000e+01)
(0.409775,7.997000e+01)
(0.413001,7.997000e+01)
(0.416228,7.997000e+01)
(0.419454,7.997000e+01)
(0.422681,7.997000e+01)
(0.425908,7.997000e+01)
(0.429134,7.997000e+01)
(0.432361,7.997000e+01)
(0.435587,7.998000e+01)
(0.438814,7.998000e+01)
(0.442040,7.998000e+01)
(0.445267,7.999000e+01)
(0.448494,7.999000e+01)
(0.451720,7.999000e+01)
(0.454947,7.999000e+01)
(0.458173,7.999000e+01)
(0.461400,7.999000e+01)
(0.464626,7.999000e+01)
(0.467853,80)
(0.471080,80)
(0.474306,80)
(0.477533,80)
(0.480759,80)
(0.483986,80)
(0.487212,80)
(0.490439,80)
(0.493666,80)
(0.496892,80)
(0.500119,80)
(0.503345,80)
(0.506572,80)
(0.509798,80)
(0.513025,8.003000e+01)
(0.516252,8.003000e+01)
(0.519478,8.003000e+01)
(0.522705,8.003000e+01)
(0.525931,8.003000e+01)
(0.529158,8.003000e+01)
(0.532384,8.003000e+01)
(0.535611,8.003000e+01)
(0.538838,8.003000e+01)
(0.542064,8.003000e+01)
(0.545291,8.003000e+01)
(0.548517,8.004000e+01)
(0.551744,8.004000e+01)
(0.554970,8.005000e+01)
(0.558197,8.005000e+01)
(0.561424,8.005000e+01)
(0.564650,8.005000e+01)
(0.567877,8.005000e+01)
(0.571103,8.006000e+01)
(0.574330,8.006000e+01)
(0.577556,8.006000e+01)
(0.580783,8.006000e+01)
(0.584010,8.006000e+01)
(0.587236,8.006000e+01)
(0.590463,8.008000e+01)
(0.593689,8.008000e+01)
(0.596916,8.010000e+01)
(0.600142,8.010000e+01)
(0.603369,8.011000e+01)
(0.606596,8.011000e+01)
(0.609822,8.011000e+01)
(0.613049,8.011000e+01)
(0.616275,8.011000e+01)
(0.619502,8.011000e+01)
(0.622728,8.011000e+01)
(0.625955,8.013000e+01)
(0.629182,8.013000e+01)
(0.632408,8.013000e+01)
(0.635635,8.013000e+01)
(0.638861,8.013000e+01)
(0.642088,8.013000e+01)
(0.645314,8.013000e+01)
(0.648541,8.013000e+01)
(0.651768,8.013000e+01)
(0.654994,8.013000e+01)
(0.658221,8.013000e+01)
(0.661447,8.014000e+01)
(0.664674,8.014000e+01)
(0.667900,8.014000e+01)
(0.671127,8.014000e+01)
(0.674354,8.015000e+01)
(0.677580,8.015000e+01)
(0.680807,8.015000e+01)
(0.684033,8.015000e+01)
(0.687260,8.016000e+01)
(0.690486,8.016000e+01)
(0.693713,8.016000e+01)
(0.696940,8.016000e+01)
(0.700166,8.017000e+01)
(0.703393,8.017000e+01)
(0.706619,8.017000e+01)
(0.709846,8.017000e+01)
(0.713072,8.017000e+01)
(0.716299,8.017000e+01)
(0.719526,8.017000e+01)
(0.722752,8.017000e+01)
(0.725979,8.017000e+01)
(0.729205,8.019000e+01)
(0.732432,8.019000e+01)
(0.735658,8.019000e+01)
(0.738885,8.019000e+01)
(0.742112,8.019000e+01)
(0.745338,8.019000e+01)
(0.748565,8.019000e+01)
(0.751791,8.019000e+01)
(0.755018,8.019000e+01)
(0.758244,8.019000e+01)
(0.761471,8.019000e+01)
(0.764698,8.019000e+01)
(0.767924,8.019000e+01)
(0.771151,8.019000e+01)
(0.774377,8.019000e+01)
(0.777604,8.019000e+01)
(0.780830,8.020000e+01)
(0.784057,8.020000e+01)
(0.787284,8.020000e+01)
(0.790510,8.020000e+01)
(0.793737,8.020000e+01)
(0.796963,8.020000e+01)
(0.800190,8.020000e+01)
(0.803416,8.021000e+01)
(0.806643,8.021000e+01)
(0.809870,8.021000e+01)
(0.813096,8.021000e+01)
(0.816323,8.021000e+01)
(0.819549,8.021000e+01)
(0.822776,8.021000e+01)
(0.826002,8.021000e+01)
(0.829229,8.021000e+01)
(0.832456,8.021000e+01)
(0.835682,8.021000e+01)
(0.838909,8.021000e+01)
(0.842135,8.023000e+01)
(0.845362,8.023000e+01)
(0.848588,8.023000e+01)
(0.851815,8.023000e+01)
(0.855042,8.023000e+01)
(0.858268,8.023000e+01)
(0.861495,8.023000e+01)
(0.864721,8.023000e+01)
(0.867948,8.023000e+01)
(0.871174,8.023000e+01)
(0.874401,8.023000e+01)
(0.877628,8.023000e+01)
(0.880854,8.024000e+01)
(0.884081,8.024000e+01)
(0.887307,8.024000e+01)
(0.890534,8.024000e+01)
(0.893760,8.024000e+01)
(0.896987,8.024000e+01)
(0.900214,8.024000e+01)
(0.903440,8.024000e+01)
(0.906667,8.024000e+01)
(0.909893,8.024000e+01)
(0.913120,8.024000e+01)
(0.916346,8.024000e+01)
(0.919573,8.024000e+01)
(0.922800,8.024000e+01)
(0.926026,8.024000e+01)
(0.929253,8.024000e+01)
(0.932479,8.024000e+01)
(0.935706,8.024000e+01)
(0.938932,8.024000e+01)
(0.942159,8.024000e+01)
(0.945386,8.024000e+01)
(0.948612,8.024000e+01)
(0.951839,8.024000e+01)
(0.955065,8.025000e+01)
(0.958292,8.025000e+01)
(0.961518,8.025000e+01)
(0.964745,8.025000e+01)
(0.967972,8.025000e+01)
(0.971198,8.025000e+01)
(0.974425,8.025000e+01)
(0.977651,8.025000e+01)
(0.980878,8.025000e+01)
(0.984104,8.025000e+01)
(0.987331,8.025000e+01)
(0.990558,8.025000e+01)
(0.993784,8.025000e+01)
(0.997011,8.025000e+01)
(1.000237,8.025000e+01)
    };
    \addplot[\mythickness,color=dingcolor, dashed] coordinates {
(0.000000,0)
(1.136539,1.465000e+01)%
    };
    \addplot[\mythickness,color=kukelovacolor, densely dotted] coordinates {
(0.000000,0)
(0.016903,6.051000e+01)
(0.033806,8.043000e+01)
(0.050709,8.596000e+01)
(0.067612,9.073000e+01)
(0.084515,9.413000e+01)
(0.101418,9.751000e+01)
(0.118322,9.907000e+01)
(0.135225,1.000100e+02)
(0.152128,1.006700e+02)
(0.169031,1.013300e+02)
(0.185934,1.023900e+02)
(0.202837,1.024100e+02)
(0.219740,1.027300e+02)
(0.236643,1.032100e+02)
(0.253546,1.033800e+02)
(0.270449,1.035200e+02)
(0.287352,1.035900e+02)
(0.304255,1.036300e+02)
(0.321159,1.038000e+02)
(0.338062,1.040700e+02)
(0.354965,1.040700e+02)
(0.371868,1.042200e+02)
(0.388771,1.043000e+02)
(0.405674,1.043300e+02)
(0.422577,1.043300e+02)
(0.439480,1.043400e+02)
(0.456383,1.043600e+02)
(0.473286,1.043800e+02)
(0.490189,1.043900e+02)
(0.507092,1.044300e+02)
(0.523995,1.045000e+02)
(0.540899,1.045100e+02)
(0.557802,1.045300e+02)
(0.574705,1.045900e+02)
(0.591608,1.045900e+02)
(0.608511,1.046200e+02)
(0.625414,1.046200e+02)
(0.642317,1.046300e+02)
(0.659220,1.046400e+02)
(0.676123,1.046400e+02)
(0.693026,1.046400e+02)
(0.709929,1.046400e+02)
(0.726832,1.046400e+02)
(0.743736,1.046400e+02)
(0.760639,1.046400e+02)
(0.777542,1.046600e+02)
(0.794445,1.046600e+02)
(0.811348,1.046600e+02)
(0.828251,1.046700e+02)
(0.845154,1.046700e+02)
(0.862057,1.046700e+02)
(0.878960,1.046700e+02)
(0.895863,1.047000e+02)
(0.912766,1.047000e+02)
(0.929669,1.047100e+02)
(0.946572,1.047100e+02)
(0.963476,1.047100e+02)
(0.980379,1.047100e+02)
(0.997282,1.047100e+02)
(1.014185,1.047100e+02)
};
\end{axis}
\end{tikzpicture}
\includestandalone[width=0.37\textwidth]{graphs/ransac/ransac_legend}
\caption{Number er of inliers vs. time.}
\label{fig:ransac}
\end{figure}

Finally, given
\begin{equation}
    \Hy =
    \begin{bmatrix}
        \hat{h}_1 & \hat{h}_2 & \hat{h}_3 \\
        \hat{h}_4 & \hat{h}_5 & \hat{h}_6 \\
        \hat{h}_7 & \hat{h}_8 & \hat{h}_9 \\
    \end{bmatrix},
\end{equation}
and comparing to~\eqref{eq:Hyparam}, we obtain the four equations
\begin{equation}
    \hat{h}_1 - \hat{h}_9 = 0,\quad \hat{h}_3 + \hat{h}_7 = 0,\quad \hat{h}_4 = \hat{h}_6 = 0\;.
\end{equation}
In conclusion, we get four cubic equations in four unknowns. Using the automatic generator,
and carefully selecting a basis~\cite{larsson2018cvpr},
we are able to find an elimination template of $10\times 15$, with five solutions,
which are obtained by solving a $5\times 5$ eigenvalue problem.

\def\droneimgscaleTwo{0.98\textwidth}
\begin{figure*}[t]
\centering
\begin{tikzpicture}
\begin{axis}[
    xlabel=$x$,
    ylabel=$z$,
    zlabel=$y$,
    z dir=reverse,
    unit vector ratio*=1 1 3,
    xmin = -2,
    xmax = 2,
    ymin = -1,
    ymax = 4,
    zmin = 0,
    zmax = 1.5,
]
\addplot3[inliercolor, mark=*, only marks, mark size=0.4pt] table {data/indoor_our_Xin.txt};
\addplot3[outliercolor, mark=*, only marks, mark size=0.4pt] table {data/indoor_our_Xout.txt};
\addplot3[gtcolor, thick, mark = none] table {data/indoor_gt_traj.txt};
\addplot3[ourcolor, mark = none, \thicknesslineplotThree] table {data/indoor_our_traj.txt};
\end{axis}
\end{tikzpicture}

\begin{tikzpicture}
\begin{axis}[
    xlabel=$x$,
    ylabel=$z$,
    zlabel=$y$,
    z dir=reverse,
    unit vector ratio*=1 1 3,
    xmin = -2,
    xmax = 2,
    ymin = -1,
    ymax = 4,
    zmin = 0,
    zmax = 1.5,
]
\addplot3[inliercolor, mark=*, only marks, mark size=0.4pt] table {data/indoor_ding_Xin.txt};
\addplot3[outliercolor, mark=*, only marks, mark size=0.4pt] table {data/indoor_ding_Xout.txt};
\addplot3[gtcolor, thick, mark = none] table {data/indoor_gt_traj.txt};
\addplot3[dingcolor, mark = none, \thicknesslineplotThree] table {data/indoor_ding_traj.txt};
\end{axis}
\end{tikzpicture}

\begin{tikzpicture}
\begin{axis}[
    xlabel=$x$,
    ylabel=$z$,
    zlabel=$y$,
    z dir=reverse,
    unit vector ratio*=1 1 3,
    xmin = -2,
    xmax = 2,
    ymin = -1,
    ymax = 4,
    zmin = 0,
    zmax = 1.5,
]
\addplot3[inliercolor, mark=*, only marks, mark size=0.4pt] table {data/indoor_kukelova_Xin.txt};
\addplot3[outliercolor, mark=*, only marks, mark size=0.4pt] table {data/indoor_kukelova_Xout.txt};
\addplot3[gtcolor, thick, mark = none] table {data/indoor_gt_traj.txt};
\addplot3[kukelovacolor, mark = none, \thicknesslineplotThree] table {data/indoor_kukelova_traj.txt};
\end{axis}
\end{tikzpicture}%
\caption{Drone experiment ground truth trajectories (black) for the \emph{indoor} sequence,
and the estimated trajectories for the different solvers. Green dots indicate inliers
(from at least one frame) and red dots denote outliers (all frames).
\emph{Left to right:} Our, Ding~\etal{}~\cite{ding-etal-2019-iccv} and
Kukelova~\etal{}~\cite{kukelova-etal-2017-cvpr}.}
\label{fig:drones2}
\end{figure*}
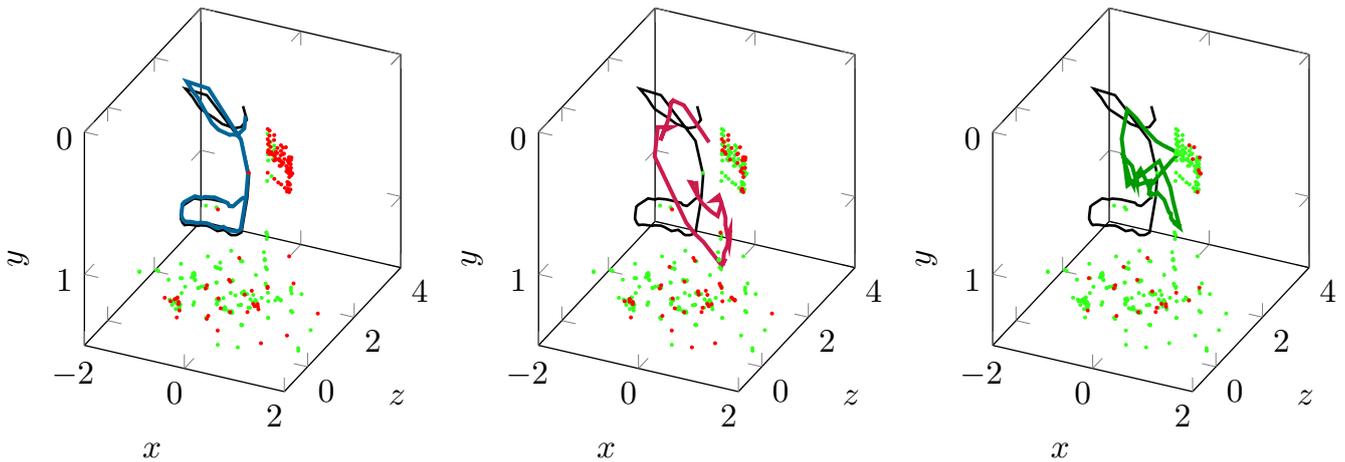

\section{Experiments}
\subsection{Synthetic experiments}
\subsubsection{Numerical stability and noise sensitivity}
In this section the proposed solvers are tested on synthetic data.
We generate points on the ground plane~$y=0$ and compare the proposed solvers
with the corresponding solvers by Ding~\etal{}~\cite{ding-etal-2019-iccv}\footnote{Code
shared by the authors.}. We only compare these
solvers as they both estimate homographies, as compared to the methods by
Kukelova~\etal{}~\cite{kukelova-etal-2017-cvpr} and Bougnoux~\cite{bougnoux-iccv-1998},
which estimates the fundamental matrix. Furthermore, with noise-free data, the latter methods
degenerate for planar correspondences.

We generate all solvers in C++, and measure the error for the obtained homography and the focal
length(s) over 10,000 randomly generated problem instances. The problem instances are obtained by
generating scene points $[x_i,\,0,\,z_i,\,1]^\T$, where $x_i,\,z_i$ are chosen from a random
distribution with zero mean and unit variance, then projected through
cameras~$\mat{P}_i=[\mat{R}_i\,|\,\mat{t}_i]$, where $\mat{R}_i=\mat{R}_{\mathrm{imu}}^{(i)}\mat{R}_z^{(i)}$.
The translation component was generated at random with zero mean and unit variance.
When measuring the error, the homographies are normalized to have last
entry equal to~1. To increase numerical stability, we use the normalization technique proposed by
Ding~\etal{}~\cite{ding-etal-2019-iccv}. The results are shown in~\cref{fig:exp1}.
For the partially calibrated cases with unknown and equal focal length ($fHf$),
and one unknown focal length ($Hf$) the proposed solvers are significantly more stable. Furthermore,
in the case of unknown and (possibly) different focal lengths ($f_1Hf_2$) the method by
Ding~\etal{}~\cite{ding-etal-2019-iccv} uses 4~point correspondences, hence the homography
estimation is a regular 4-point DLT system obtained using SVD, with known stable properties.

To mimic real data, we distort the image point correspondences by adding zero mean Gaussian noise
with a standard deviation~$\sigma_N$, and vary the noise level.
To be able to compare the same quantity, the obtained homographies and fundamental matrices,
are decomposed to
relative translation and orientation.
We  proceed as in the previous experiment by generating ground truth data, but also include
non-planar point correspondences for the 6-point and 7-point methods, in order for them to not
degenerate.

We define the errors as in~\cite{saurer-etal-2017,ding-etal-2019-iccv},
namely
\begin{equation}\label{eq:errors}
\begin{aligned}
    e_{\mat{R}} &= \arccos\!\left(\fr{\tr(\mat{R}_{\mathrm{GT}} \mat{R}_{\mathrm{est}}^\T) - 1}{2}\right), \\
    e_{\mat{t}} &= \arccos\!\left(\fr{\mat{t}_{\mathrm{GT}}^\T\mat{t}_{\mathrm{est}} }{\norm{\mat{t}_{\mathrm{GT}}} \norm{\mat{t}_{\mathrm{est}}}}\right), \\
    e_{f} &= \fr{|f_{\mathrm{GT}} - f_{\mathrm{est}}|}{f_{\mathrm{GT}}}.
\end{aligned}
\end{equation}

In \cref{fig:noise} we show the results for the noise sensitivity experiments. We note
that the proposed solvers perform consistent to SOTA solvers in all cases. We emphasize,
however, that, for the synthetic data experiments, the main benefit of using our method
is the execution time, which we show in the next section.

\subsubsection{Speed evaluation}
The experiments were run on a laptop with an Intel Core i5-6200U 2.30GHz CPU using
C++ implementations based on the Eigen linear algebra library~\cite{eigen}. The same optimization
flags were used for all solvers. The mean runtime for one hypothesis estimation is shown
in \cref{tab:exectime}. We note a speed-up of more than 75$\times$ for our $fHf$ solver compared
to the SOTA solver by Ding~\etal{}~\cite{ding-etal-2019-iccv}. Furthermore, we argue
that this is the most relevant case for indoor UAV navigation, as most consumer-grade UAVs
have fixed focal length; however, when considering more than one drone,
the other cases are relevant. For the case of $Hf$, one could have a calibrated
camera on one of the UAVs, as a reference, thus not having to calibrate
all remaining UAVs, potentially eliminating exhaustive calibration procedures.
For the $f_1Hf_2$ case we may consider two drones (with possibly different focal
lengths) covering the same area,~\eg{} drone swarms~\cite{schilling-etal-2019}, or drones
with varifocal optics.

Apart from being significantly faster than the SOTA solvers, our approach benefits from requiring
one less point correspondence. In practice, when one uses a robust framework, such as RANSAC,
the number of iterations required depends on the number of points to select, hence the
practical speed-up is greater than 75$\times$, see~\cref{tab:exectime2}.

Furthermore, the 2.5-point and 3.5-point methods both benefit from being able to do
a consistency check of the putative solutions, by using the third or fourth point correspondence, respectively, whereas
other methods need to estimate the hypothesis every iteration. This optimization step
was implemented for the $fHf$ solvers, and we compare the number of inliers vs total execution
time, see~\cref{fig:ransac}, in a complete RANSAC framework.
In this experiment 100 points were generated on the ground plane, as well as 30~point non-planar
correspondences, in order to simulate a possible real-life scenario. Noise was added, and 20~\% of the
points were scrambled to simulate outliers.
As is evident, our method is faster than both methods we compare to, and significantly faster
than the SOTA solver~\cite{ding-etal-2019-iccv}.

\begin{table}[htb]
\centering
\caption{Mean execution time for 10,000 randomly generated problems in C++.
All solvers were implemented using Eigen~\cite{eigen} and compiled in g++
with the \texttt{-O3} optimization flag.}
\vspace{-0.1cm}
\begin{tabular}{lll}
\hline
Case      & Author                                          & Execution time ($\mu$s)\\ \hline
$fHf$     & Our                                             & 14    \\
          & Ding~\etal{}~\cite{ding-etal-2019-iccv}         & 1052  \\
$fEf$     & Kukelova~\etal{}~\cite{kukelova-etal-2017-cvpr} & 103   \\ \hline
$Hf$      & Our                                             & 5     \\
          & Ding~\etal{}~\cite{ding-etal-2019-iccv}         & 124   \\
$Ef$      & Kukelova~\etal{}~\cite{kukelova-etal-2017-cvpr} & 25    \\ \hline
$f_1Hf_2$ & Our                                             & 9     \\
          & Ding~\etal{}.~\cite{ding-etal-2019-iccv}        & 47    \\
$f_1Ef_2$ & Bougnoux~\cite{bougnoux-iccv-1998}              & 27    \\ \hline
\end{tabular}
\label{tab:exectime}
\end{table}

\begin{table}[htb]
\centering
\caption{Mean execution time for reaching at least 95 \% inlier ratio in a RANSAC loop,
based on 100 randomly generated problems in C++.}
\vspace{-0.1cm}
\begin{tabular}{lll}
\hline
Case      & Author                                          & Execution time (ms)\\ \hline
$fHf$     & Our                                             & 0.035    \\
          & Ding~\etal{}~\cite{ding-etal-2019-iccv}         & 19.32  \\
$fEf$     & Kukelova~\etal{}~\cite{kukelova-etal-2017-cvpr} & 0.118   \\ \hline
\end{tabular}
\label{tab:exectime2}
\end{table}

\def\droneimgscaleOne{0.228\textwidth}
\def\droneimgscale{0.16\textwidth}
\def\realexpheight{0.32\textwidth}

\begin{figure*}[t]
\centering
\begin{tabular}{l l c c}
Basement (610 frames, 2978 keypoints) \hspace*{-25mm} \\
\raisebox{5mm}{\includegraphics[width=\droneimgscale]{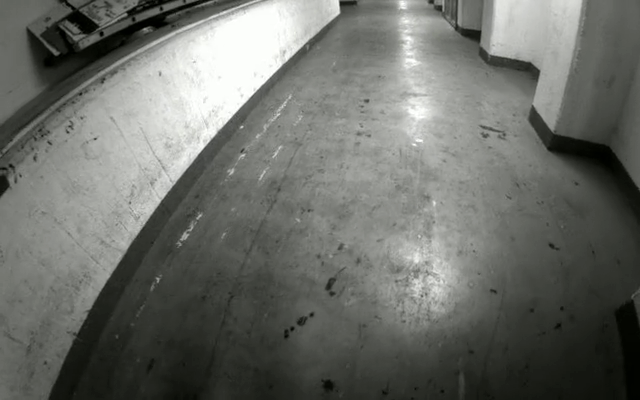}}&%
\includestandalone[width=\droneimgscaleOne]{graphs/real_data_error/basement_f_error}&%
\includestandalone[width=\droneimgscaleOne]{graphs/real_data_error/basement_R_error}&%
\includestandalone[width=\droneimgscaleOne]{graphs/real_data_error/basement_t_error}\\%
Carpet (107 frames, 400 keypoints) \hspace*{-25mm} \\
\raisebox{5mm}{\includegraphics[width=\droneimgscale]{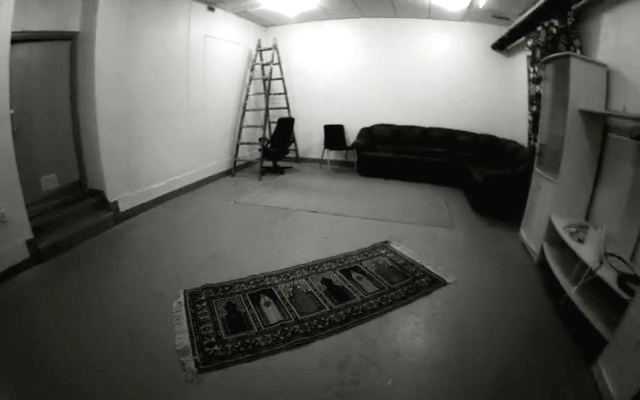}}&%
\includestandalone[width=\droneimgscaleOne]{graphs/real_data_error/carpet_f_error}&%
\includestandalone[width=\droneimgscaleOne]{graphs/real_data_error/carpet_R_error}&%
\includestandalone[width=\droneimgscaleOne]{graphs/real_data_error/carpet_t_error}\\%
Indoor (48 frames, 184 keypoints) \hspace*{-25mm} \\
\raisebox{5mm}{\includegraphics[width=\droneimgscale]{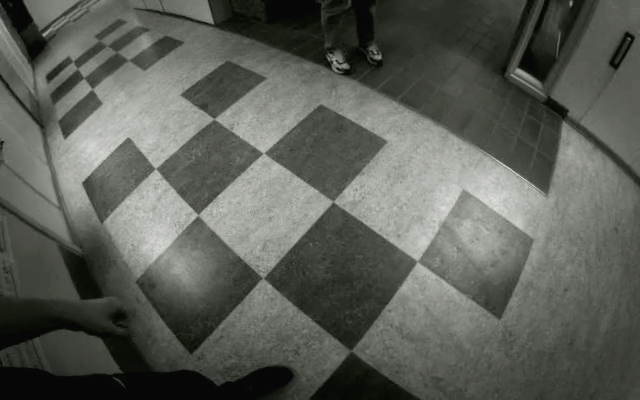}}&%
\begin{tikzpicture}[>=latex]
    \begin{axis}[
  width=0.5\textwidth,
  height=\realexpheight,
  axis x line*=bottom,
  axis y line*=left,
  ylabel={Focal length error} ,
  ymin = -4.1,
  ymax = 0.1,
  xmin = 1,
  xmax = 47,
  ytick = {-4, -2, 0, 2},
  yticklabels={$10^{-4}$,$10^{-2}$, 1, $10^2$},
  xlabel=Frame No.,
  ]
\addplot[\thicknesslineplotTwo, color=ourcolor] coordinates {
    (1, -1.499867)
    (2, -0.973445)
    (3, -1.603290)
    (4, -1.118631)
    (5, -1.357896)
    (6, -0.427107)
    (7, -1.010965)
    (8, -1.541421)
    (9, -2.298233)
    (10, -1.565756)
    (11, -1.502944)
    (12, -1.879330)
    (13, -1.287044)
    (14, -1.945950)
    (15, -1.949471)
    (16, -1.077880)
    (17, -3.797503)
    (18, -1.236461)
    (19, -0.907829)
    (20, -1.068892)
    (21, -0.546582)
    (22, -1.251035)
    (23, -1.383391)
    (24, -1.389038)
    (25, -0.551983)
    (26, -1.540481)
    (27, -1.731128)
    (28, -1.145867)
    (29, -1.315418)
    (30, -1.959136)
    (31, -1.395823)
    (32, -1.144698)
    (33, -1.940726)
    (34, -1.816207)
    (35, -1.088439)
    (36, -1.245196)
    (37, -1.809992)
    (38, -0.430100)
    (39, -0.539860)
    (40, -1.141010)
    (41, -3.298293)
    (42, -1.674695)
    (43, -1.505198)
    (44, -0.702575)
    (45, -1.303000)
    (46, -1.759848)
    (47, -2.485388)
};

\addplot[\thicknesslineplotTwo, color=dingcolor] coordinates {
    (1, -1.430896)
    (2, -1.298068)
    (3, -0.281418)
    (4, -1.435377)
    (5, -0.957512)
    (6, -0.898584)
    (7, -1.081302)
    (8, -0.716708)
    (9, -1.374537)
    (10, -0.754134)
    (11, -0.719119)
    (12, -0.700273)
    (13, -1.526102)
    (14, -0.179651)
    (15, -1.065930)
    (16, -2.187159)
    (17, -1.336881)
    (18, -0.211670)
    (19, -2.171272)
    (20, -0.527319)
    (21, -1.086769)
    (22, -1.741768)
    (23, -0.190487)
    (24, -0.976524)
    (25, -0.333505)
    (26, -1.265413)
    (27, -2.339893)
    (28, -1.147329)
    (29, -1.256020)
    (30, -0.950328)
    (31, -1.135508)
    (32, -0.168031)
    (33, -1.659046)
    (34, -0.222448)
    (35, -0.906344)
    (36, -0.865546)
    (37, -0.320760)
    (38, -0.312812)
    (39, -0.343586)
    (40, -0.252882)
    (41, -1.122982)
    (42, -0.954635)
    (43, -0.253283)
    (44, -0.684268)
    (45, -1.384535)
    (46, -1.384744)
    (47, -0.518804)
};

\addplot[\thicknesslineplotTwo, color=kukelovacolor] coordinates {
    (1, -0.244043)
    (2, -0.637980)
    (3, -0.736533)
    (4, -0.517034)
    (5, -0.353754)
    (6, -0.690787)
    (7, -0.344034)
    (8, -0.372264)
    (9, -0.286710)
    (10, -0.210496)
    (11, -0.193261)
    (12, -0.817069)
    (13, -0.501311)
    (14, -0.235468)
    (15, -0.567036)
    (16, -0.395616)
    (17, -0.270694)
    (18, -0.228123)
    (19, -0.557143)
    (20, -0.502272)
    (21, -0.173432)
    (22, -0.429137)
    (23, -0.557426)
    (24, -0.283905)
    (25, -0.420943)
    (26, -0.529517)
    (27, -0.805066)
    (28, -0.273778)
    (29, -0.438876)
    (30, -0.500949)
    (31, -1.095665)
    (32, -1.043256)
    (33, -1.240211)
    (34, -0.522438)
    (35, -0.292312)
    (36, -1.238428)
    (37, -3.113972)
    (38, -0.721977)
    (39, -0.261558)
    (40, -0.599519)
    (41, -0.225062)
    (42, -0.952431)
    (43, -1.173499)
    (44, -0.910887)
    (45, -0.444168)
    (46, -1.751618)
    (47, -0.875629)
};
\end{axis}
\end{tikzpicture}&%
\begin{tikzpicture}[>=latex]
    \begin{axis}[
  width=0.5\textwidth,
  height=\realexpheight,
  axis x line*=bottom,
  axis y line*=left,
  ylabel={Rotation error (deg)} ,
  ymin = -4.1,
  ymax = 2.1,
  xmin = 1,
  xmax = 47,
  ytick = {-4, -2, 0, 2},
  yticklabels={$10^{-4}$,$10^{-2}$, 1, $10^2$},
  xlabel=Frame No.,
  ]
\addplot[\thicknesslineplotTwo, color=ourcolor] coordinates {
    (1, -1.043891)
    (2, -1.502228)
    (3, -1.341782)
    (4, -1.103399)
    (5, -0.838160)
    (6, -1.192165)
    (7, -1.325979)
    (8, -0.657702)
    (9, -0.822703)
    (10, -0.763524)
    (11, -1.366010)
    (12, -0.534719)
    (13, -0.825053)
    (14, -0.745930)
    (15, -2.010196)
    (16, -0.400910)
    (17, -0.249478)
    (18, -0.635028)
    (19, -0.277220)
    (20, -2.110270)
    (21, -1.087817)
    (22, -0.679720)
    (23, -0.911231)
    (24, -0.660388)
    (25, -1.009578)
    (26, -0.517324)
    (27, -1.392299)
    (28, -0.573551)
    (29, -0.380457)
    (30, -0.590770)
    (31, -0.585074)
    (32, -0.765347)
    (33, -0.162441)
    (34, -1.786203)
    (35, -0.005089)
    (36, -0.288229)
    (37, -0.152881)
    (38, 0.636458)
    (39, 0.373167)
    (40, -0.391923)
    (41, -1.085768)
    (42, -0.646673)
    (43, -0.214252)
    (44, -0.312672)
    (45, -0.522493)
    (46, -0.361108)
    (47, -0.193463)
};

\addplot[\thicknesslineplotTwo, color=dingcolor] coordinates {
    (1, -0.216297)
    (2, -0.130266)
    (3, -0.289146)
    (4, -0.788800)
    (5, -0.299765)
    (6, -0.450148)
    (7, 0.198234)
    (8, -0.002212)
    (9, -0.342524)
    (10, 0.157759)
    (11, 0.430173)
    (12, 1.851357)
    (13, 1.982960)
    (14, 1.544350)
    (15, 0.690183)
    (16, 0.251075)
    (17, 0.593552)
    (18, 0.574404)
    (19, -0.047708)
    (20, 0.379896)
    (21, -0.051691)
    (22, 0.376707)
    (23, 0.769577)
    (24, 0.056432)
    (25, 0.341157)
    (26, -0.248214)
    (27, 0.311345)
    (28, 0.120604)
    (29, 0.426947)
    (30, 0.604757)
    (31, 1.757554)
    (32, 2.112078)
    (33, 0.391870)
    (34, 1.035989)
    (35, 1.444597)
    (36, 1.838857)
    (37, 1.854758)
    (38, 0.711611)
    (39, 0.660268)
    (40, 0.831952)
    (41, 0.428045)
    (42, 0.696538)
    (43, 0.428350)
    (44, 0.541489)
    (45, 0.388775)
    (46, 1.788713)
    (47, 1.818215)
};

\addplot[\thicknesslineplotTwo, color=kukelovacolor] coordinates {
    (1, 0.799932)
    (2, 0.051448)
    (3, 0.398080)
    (4, 0.100164)
    (5, 0.756665)
    (6, 0.016134)
    (7, 0.576942)
    (8, 0.034582)
    (9, 0.542730)
    (10, 0.428003)
    (11, 0.893157)
    (12, 0.268134)
    (13, 0.308941)
    (14, 0.727787)
    (15, 0.381454)
    (16, 0.105292)
    (17, 0.643575)
    (18, 0.184581)
    (19, 0.221512)
    (20, 0.949772)
    (21, 0.710403)
    (22, 0.618908)
    (23, 0.540722)
    (24, 1.268761)
    (25, 0.624699)
    (26, 0.767203)
    (27, 0.476656)
    (28, 0.593633)
    (29, 0.909553)
    (30, 0.379534)
    (31, -0.497260)
    (32, 0.117002)
    (33, 0.129215)
    (34, 0.985090)
    (35, 1.524965)
    (36, 0.594000)
    (37, -0.086493)
    (38, 0.492274)
    (39, 0.549343)
    (40, 0.565749)
    (41, 0.735742)
    (42, 0.763062)
    (43, 0.072042)
    (44, 0.047828)
    (45, 0.079627)
    (46, -0.702215)
    (47, 0.770027)
};
\end{axis}
\end{tikzpicture}&%
\begin{tikzpicture}[>=latex]
    \begin{axis}[
  width=0.5\textwidth,
  height=\realexpheight,
  axis x line*=bottom,
  axis y line*=left,
  ylabel={Translation error (deg)} ,
  ymin = -0.5,
  ymax = 2.1,
  xmin = 1,
  xmax = 47,
  ytick = {0, 1, 2},
  yticklabels={1, 10, $10^2$},
  xlabel=Frame No.,
  ]
\addplot[\thicknesslineplotTwo, color=ourcolor] coordinates {
    (1, 0.472101)
    (2, 0.313498)
    (3, 0.783067)
    (4, 0.751580)
    (5, 0.144068)
    (6, 0.994302)
    (7, 0.436544)
    (8, 0.626644)
    (9, -0.375955)
    (10, 0.640553)
    (11, 0.378228)
    (12, -0.488369)
    (13, 0.790878)
    (14, -0.383750)
    (15, 0.417791)
    (16, 0.486950)
    (17, 0.954858)
    (18, 0.644115)
    (19, 0.799726)
    (20, 0.051006)
    (21, 0.907275)
    (22, -0.363912)
    (23, 0.530576)
    (24, -0.058540)
    (25, 0.888142)
    (26, 0.814448)
    (27, 0.918569)
    (28, 1.080300)
    (29, 0.001779)
    (30, 0.653054)
    (31, 0.088950)
    (32, 0.258479)
    (33, 0.631424)
    (34, -0.160787)
    (35, 0.437223)
    (36, 0.056985)
    (37, 0.452982)
    (38, 1.588350)
    (39, 1.400156)
    (40, -0.071855)
    (41, 0.427481)
    (42, 0.324357)
    (43, 1.326182)
    (44, 0.851602)
    (45, 0.692958)
    (46, 0.381660)
    (47, -0.122805)
};

\addplot[\thicknesslineplotTwo, color=dingcolor] coordinates {
    (1, 0.354423)
    (2, 0.805150)
    (3, 1.140345)
    (4, 0.779375)
    (5, 0.546362)
    (6, 0.778690)
    (7, 0.028549)
    (8, 1.311886)
    (9, 0.435258)
    (10, 1.144776)
    (11, 0.947864)
    (12, 1.730974)
    (13, 1.423720)
    (14, 1.678884)
    (15, 1.335118)
    (16, 0.204563)
    (17, 1.478970)
    (18, 1.227000)
    (19, 0.866707)
    (20, 0.815158)
    (21, 1.009048)
    (22, 0.877548)
    (23, 1.872815)
    (24, 0.836166)
    (25, 1.147824)
    (26, 0.465782)
    (27, 1.507253)
    (28, 1.339165)
    (29, 1.295867)
    (30, 1.715472)
    (31, 1.769836)
    (32, 1.209847)
    (33, 1.303909)
    (34, 1.588517)
    (35, 1.440510)
    (36, 1.436173)
    (37, 1.621974)
    (38, 1.792223)
    (39, 1.743817)
    (40, 1.655931)
    (41, 1.317744)
    (42, 1.434320)
    (43, 1.389065)
    (44, 1.625085)
    (45, 1.739993)
    (46, 1.531284)
    (47, 1.850234)
};

\addplot[\thicknesslineplotTwo, color=kukelovacolor] coordinates {
    (1, 1.898484)
    (2, -0.472623)
    (3, 1.796479)
    (4, 1.709045)
    (5, 1.663052)
    (6, 1.282610)
    (7, 0.091525)
    (8, 1.568741)
    (9, 1.287093)
    (10, 1.378669)
    (11, 1.778954)
    (12, 0.267478)
    (13, 1.590251)
    (14, 1.344530)
    (15, 0.377077)
    (16, 0.611137)
    (17, 1.477329)
    (18, 1.475215)
    (19, 0.957405)
    (20, 1.654388)
    (21, 1.806076)
    (22, 1.186744)
    (23, 1.640012)
    (24, 1.942328)
    (25, 1.422649)
    (26, 1.812093)
    (27, 1.846620)
    (28, 1.917425)
    (29, 1.889962)
    (30, 1.916969)
    (31, 0.720966)
    (32, 0.475311)
    (33, 1.092845)
    (34, 1.550553)
    (35, 1.620175)
    (36, 0.735036)
    (37, 0.615344)
    (38, 1.204584)
    (39, 1.759319)
    (40, 1.157958)
    (41, 1.526294)
    (42, 1.793805)
    (43, 1.470856)
    (44, 1.088132)
    (45, 1.810080)
    (46, 0.704957)
    (47, 0.518399)
};
\end{axis}
\end{tikzpicture}\\%
Outdoor (601 frames, 3659 keypoints) \hspace*{-25mm} \\
\raisebox{5mm}{\includegraphics[width=\droneimgscale]{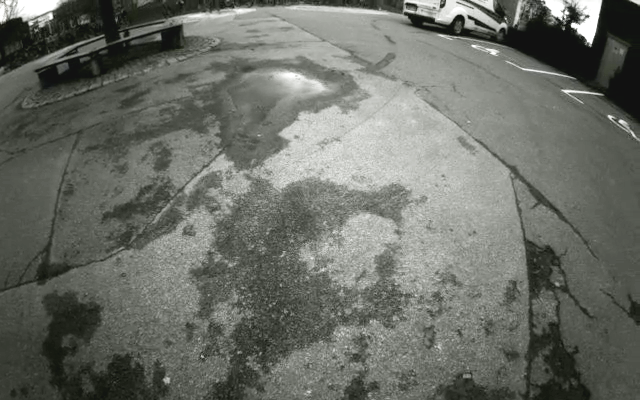}}&%
\includestandalone[width=\droneimgscaleOne]{graphs/real_data_error/outdoor_f_error}&%
\includestandalone[width=\droneimgscaleOne]{graphs/real_data_error/outdoor_R_error}&%
\includestandalone[width=\droneimgscaleOne]{graphs/real_data_error/outdoor_t_error}\\%
\end{tabular}
\includestandalone[width=0.4\textwidth]{graphs/real_data_error/real_exp_legend}\
\caption{Drone experiment: (Left to right) Example image of the scene from the UAV.
Focal length error, rotation error, translation error (as defined in~\eqref{eq:errors}).}
\label{fig:drones1}
\end{figure*}

\subsection{Real data}
The real data was captured using a monochrome global shutter camera (OV9281) with resolution $480 \times 640$ and an inertial measurement unit (MPU-9250). The extracted feature locations were undistorted to remove fish-eye effects. Ground truth and pair-wise feature matches were generated by a simultaneous localization and mapping system, where both the re-projection and IMU error were minimized; this is in order to create a globally consistent solution in metric scale. No assumptions about the scene structure were made, which lead to some matches not belonging to the ground plane, resulting in natural outliers, for the proposed method.

The data consists of both indoor and outdoor sequences containing mostly planar surfaces.
The dataset contains shorter sequences (containing fewer than 50 images), as well as longer
ones (containing more than 600 images), with varying kinds of motions.
Examples of input images can be seen in the left-most column of~\cref{fig:drones1}. Here,
we also present the focal length, rotation and translation errors compared to the ground truth
data for all frames. The estimated poses were obtained using 200 RANSAC iterations per method
and frame for the equal and unknown focal length solvers.

To visualize the qualitative difference between the methods, we show the estimated trajectories
for the~\emph{indoor} sequence in~\cref{fig:drones2}. The initial pose was synchronized with the
ground truth pose and the length of the translations scaled to match the ground truth translation
vector for each frame. From here we converted the relative poses to the absolute poses
and display the estimated trajectories together with the ground truth trajectories individually.

We also include green dots indicating feature points that have been matched (at least once)
as an inlier in the sequence. The red dots are feature points that have consistently been
rejected as outliers. For the proposed method a red cluster of rejected outliers is clearly
visible, and these are all feature point of a door, hence do not belong to the ground plane.
It is interesting to see that the method by Ding~\etal{}~\cite{ding-etal-2019-iccv} uses
these points as well when trying to estimate the relative pose. This, of course, is valid
given the more general assumption used, by allowing arbitrary plane normals; however,
it is readily seen
from the estimated trajectory alone, that this approach is not
robust enough for real-life applications.

\section{Conclusions}

In this paper we have presented new minimal solvers to perform homography based indoor positioning. The solvers can utilize IMU data to obtain a partial extrinsic calibration. We have evaluated the solvers on both synthetic and real world data, where we have shown that they are robust to both Gaussian noise, as well as noise apparent in real world data. Comparing with the current state of the art, the solvers are on par with SOTA on synthetic data and outperforms them on real world data, where the solvers can be seen to follow ground truth more accurately. However, the main benefit of the solvers compared to the current state of the art is the execution speed, which is several orders of magnitude faster. In addition to this the solvers require fewer points compared to their counterparts. These attributes make the solvers viable for real-time systems, such as UAVs, requiring many RANSAC iterations to ensure robustness and therefore fast execution to meet the short time budget per frame.

\section*{Acknowledgements}
We thank Ding~\etal{} for providing their implementations.
This work was partially supported by the Swedish Research Council
(grant no. 2015-05639),
the strategic research projects ELLIIT
and eSSENCE, the Swedish Foundation for Strategic Research project, Semantic
Mapping and Visual Navigation for Smart Robots (grant no. RIT15-0038), and
Wallenberg AI, Autonomous Systems and Software Program (WASP) funded by
Knut and Alice Wallenberg Foundation.

\bibliographystyle{IEEEtran}
\bibliography{IEEEabrv,uav}
\end{document}